\documentclass{article}

\usepackage{PRIMEarxiv}
\usepackage[utf8]{inputenc} 
\usepackage[T1]{fontenc}    
\usepackage{hyperref}       
\usepackage{url}            
\usepackage{nicefrac}       
\usepackage{microtype}      
\usepackage{lipsum}
\usepackage{fancyhdr}       
\usepackage{graphicx}
\graphicspath{{media/}}     
\usepackage[affil-it]{authblk}

\usepackage{amsmath,amssymb,amsfonts}
\usepackage{algorithmic}
\usepackage{textcomp}

\usepackage[table]{xcolor}

\usepackage{booktabs}   
\usepackage{multirow}
\usepackage{tabularx}
\usepackage{makecell}
\usepackage{rotating} 

\usepackage{enumitem}

\usepackage{silence}
\usepackage[compatibility=false]{caption}
\usepackage{chngcntr}
\usepackage{float}

\title{Consistency Analysis of Sentiment Predictions using Syntactic \& Semantic Context Assessment Summarization (SSAS)
}

\author[1]{Sharookh Daruwalla}
\author[1]{Nitin Mayande}
\author[1]{Shreeya Verma Kathuria}
\author[2]{Nitin Joglekar}
\author[3]{Charles Weber}

\affil[1]{Tellagence Inc.\thanks{\texttt{\{sharookh, nitin, shreeya\}@tellagence.com}}}
\affil[2]{Villanova School of Business, Villanova University\thanks{\texttt{nitindra.joglekar@villanova.edu}}}
\affil[3]{Maseeh College of Engineering and Computer Science, Portland State University\thanks{\texttt{webercm@pdx.edu}}}

\begin{document}
\maketitle

\begin{abstract}
The fundamental challenge of using Large Language Models (LLMs) for reliable, enterprise-grade analytics, such as sentiment prediction, is the conflict between the LLMs' inherent stochasticity (generative, non-deterministic nature) and the analytical requirement for consistency. The LLM inconsistency, coupled with the noisy nature of chaotic modern datasets, renders sentiment predictions too volatile for strategic business decisions. To resolve this, we present a Syntactic \& Semantic Context Assessment Summarization (SSAS) framework for establishing context. Context established by SSAS functions as a sophisticated data pre-processing framework that enforces a bounded attention mechanism on LLMs. It achieves this by applying a hierarchical classification structure (Themes, Stories, Clusters) and an iterative Summary-of-Summaries (SoS) based context computation architecture. This endows the raw text with high-signal, sentiment-dense prompts, that effectively mitigate both irrelevant data and analytical variance.

We empirically evaluated the efficacy of SSAS, using Gemini 2.0 Flash Lite, against a direct-LLM approach across three industry-standard datasets – Amazon Product Reviews, Google Business Reviews, Goodreads Book Reviews - and multiple robustness scenarios. Our results show that our SSAS framework is capable of significantly improving data quality,  up to 30\%, through a combination of noise removal and improvement in the estimation of sentiment prediction. Ultimately, consistency in our context-estimation capabilities provides a stable and reliable evidence base for decision-making.
\end{abstract}


\section{Introduction}
In today’s high-paced business landscape, the mandate for data-driven decision-making is absolute \cite{berger_uniting_2020}. However, a significant gap exists between the volume of available textural data and the infrastructure required to extract actionable insights from it. While strategic decisions rely on rigorous data analysis, the sheer scale of non-trivial, modern datasets creates a chaotic environment where signal is frequently buried under layers of technical friction. This challenge is further exacerbated by the integration of Large Language Models (LLMs) \cite{manning_introduction_2008, brown_language_2020, chen_can_2024, zhao_calibrate_2021, dong_survey_2024, srivastava_beyond_2023, agrawal_large_2022, bang_multitask_2023} whose probabilistic architecture is fundamentally at odds with the repeatable and precise output requirements of enterprise-grade reporting. The conflict between operational velocity and analytical precision is defined by the following systemic pressures:
\begin{itemize}
    \item The Velocity Mandate: Competitive advantage is predicated on the ability to process large-scale datasets quickly. Any latency in the pipeline results in decayed relevance.
    \item The Signal-to-Noise Deficit: Modern large-scale text processing data (e.g., social media marketing datasets) are fundamentally chaotic. While a fraction of the data is focused on the core problem, the vast majority is comprised of noisy, inconsequential information that complicates processing.
    \item The Strategic Skill Gap: Large-scale text-processing teams are domain experts e.g., social media marketers are experts in brand strategy and consumer behavior. These experts ought not be expected to possess the specialized engineering skills required to manipulate stochastic LLM outputs into reliable data.
\end{itemize}

This environment presents two primary hurdles: the Noise problem i.e. the difficulty of identifying, isolating, and removing irrelevant data, and the Inconsistency problem which is the inability to ensure that an LLM-based analytical process is consistent. While these hurdles remain, teams are left with analytics, such as sentiment predictions \cite{deng_llms_2023, maas_learning_2011, mohammad_sentiment_2021, niu_review_2021, rosenthal_columbia_2014, sharma_movie_2023}, that are too volatile to support enterprise-level decision-making. The genesis of these problems lie in utilizing tools, e.g. LLM models, that are currently designed for generative creativity rather than consistent data processing.

\subsection{The Paradox of LLM Creativity: Why Generative AI underperforms in Data Science}
To bridge the gap between AI potential and large-scale text processing execution, we must address the fundamental conflict between LLM architectures and requirements of data science. LLMs are, by design, engines of probability. While their underlying mechanics are revolutionary for creative synthesis, they are inherently poorly suited for the rigid, invariant requirements of data analytics. The root of this failure lies in the LLM’s attention mechanism \cite{niu_review_2021}. In a standard generative configuration, the attention mechanism dictates which tokens the model prioritizes during processing \cite{vaswani_attention_2023}. Because these models are optimized for novelty, the mechanism may assign different weights to the same input tokens across successive runs \cite{bender_dangers_2021}. This stochasticity is an asset for creative tasks, but it can be a significant liability for data science tasks where latent space stability is required to ensure data integrity. In sentiment prediction \cite{mohammad_sentiment_2021, rosenthal_columbia_2014, sharma_movie_2023, sun_sentiment_2023, wang_aligning_2023, zhang_sentiment_2023}, this creative variance manifests as inconsistency \cite{chen_can_2024, bang_multitask_2023, niu_review_2021, shi_large_2023}. If the same dataset yields different sentiment scores upon re-run, the output is functionally useless for strategic planning. At a strategic level, this inconsistency is more than a technical glitch; it is an erosion of the evidentiary basis for decision-making. To achieve reliability, we must move from generative variance toward a consistent standard. 

\subsection{Requisite Capability: Consistency in Summarization and Sentiment Prediction}
For AI-driven sentiment analysis \cite{mohammad_sentiment_2021, rosenthal_columbia_2014, sharma_movie_2023, sun_sentiment_2023, wang_aligning_2023, zhang_sentiment_2023} to meet the threshold of enterprise-grade analytics, it must adhere to the benchmark of consistency. This is not a subjective measure of quality but a technical requirement for any system intended to serve as a stable foundation for corporate strategy. Consistency is defined as the ability to generate an identical output when provided with the same input. In a professional analytical environment, an analysis performed today must be perfectly replicable tomorrow. Without this guarantee, data-driven insights are merely ephemeral snapshots, lacking the stability required for long-term strategic investments.

Achieving this standard necessitates a rigorous approach to identifying and isolating noise. Research suggests that the inclusion of irrelevant information within the  input can be damaging to performance as it forces the model to attend to inconsequential patterns \cite{shi_large_2023}. This creates a signal-to-noise deficit that is particularly acute in modern marketing and commercial datasets, where the sheer volume of data often buries actionable insights under layers of technical friction \cite{apostolova_training_2018}. A consistent analytical framework must, therefore, be capable of (1) identifying the relevancy of data points within the broader dataset, (2) isolating core problem-related data from inconsequential information, and finally (3) removing noise to ensure attention focused exclusively on relevant context \cite{niu_review_2021, shi_large_2023}. By leveraging these capabilities, organizations can transform LLMs from creative assistants to precise analytical instruments for meeting the demands of large-scale text processing.

\subsection{Hierarchical Contextual Framework for Analytical Integrity}
Our SSAS framework \cite{mayande_syntactic_2024, mayande_leveraging_2025, kathuria_leveraging_2026} posits a method to address the dual crises of noise and stochastic inconsistency by replacing the black box unpredictability of standard LLMs with a structured methodology designed to enforce integrity on chaotic datasets \cite{liu_pre-train_2023}. The methodology is implemented through a specialized two-phase framework:
\begin{enumerate}
    \item Contextual Relevancy: The process begins by evaluating the data within its specific context. By identifying information relevancy at a granular level, the system determines which data points are pertinent to the defined problem and which are extraneous.
    \item Noise Reduction and Reliability Improvement: This is the critical point of interdependence. Using the derived context from Phase 1, we systematically reduce dataset noise. By feeding only the refined, relevant context into the LLM, we neutralize variance and significantly improve the consistency and reliability of the output.
\end{enumerate}

Our framework refines raw, chaotic data into a reliable  and analytically relevant dataset. In parallel, by narrowing the model's focus through derived context, this framework ensures that the refined input consistently yields the same results, addressing, to a large extent, the stochasticity problem inherent in generative architectures. This approach alleviates the technical burden on large-scale text processing teams, allowing them to focus on strategy rather than engineering methods \cite{davenport_all-ai_2023}. Our contributions in this paper are as follows.
\begin{enumerate}
    \item We present our SSAS framework to provide LLMs with assisting context which, in turn, helps LLMs to focus their attention mechanism and provide consistency to analytical tasks.
    \item Our framework classifies datasets into a hierarchical structure of Themes, Stories, and Clusters to create consistent summaries and Summaries-of-Summaries (SoS) across multiple levels of data aggregations.
    \item Our framework helps identify noisy data points within datasets and helps LLMs focus their attention on the signal within the datasets.
    \item Our framework is able to significantly improve data quality, up to 30\%, through a combination of noise removal and improvement in the estimation of sentiment prediction. 
\end{enumerate}

The rest of this paper is structured as follows: Section 2 presents related and background work around LLM mechanisms, hierarchical information, and semantic alignment. Section 3 presents our Syntactic \& Semantic Context Assessment Summarization (SSAS) framework. Section 4 provides details on our Evaluation Framework, while Section 5 presents the results of our framework when compared against a Direct-LLM approach. We conclude in Section 6.

\section{Related Works}
The emergence of Large Language Models (LLMs) has fundamentally redefined text analysis, shifting the paradigm from supervised feature engineering toward zero-shot and few-shot learning \cite{brown_language_2020, ma_digeo_2023}. However, as these models move from creative synthesis to enterprise-grade analytics, their inherent instability presents significant challenges. Our work builds upon three primary areas of research: the sensitivity of in-context learning \cite{min_metaicl_2022}, the mechanics of attention-based noise \cite{niu_review_2021, shi_large_2023}, and hierarchical data summarization. Elements of sections 2 and 3 are similar to Kathuria et al.\cite{kathuria_leveraging_2026}, because our respective works share and build on the SSAS framework originally laid out in Mayande et al.\cite{mayande_syntactic_2024}. For ease of access, we have summarized these ideas in this paper, so that readers need not go back to Kathuria et al. (op. cite).

\subsection{In-Context Learning and Prompt Instability}
The efficacy of Large Language Models (LLMs) in zero-shot and few-shot regimes is largely governed by the paradigm of In-Context Learning (ICL). However, despite their sophisticated semantic latent spaces, LLMs exhibit a profound and notorious sensitivity to the specificities of the input context. Zhao et al. \cite{zhao_calibrate_2021} characterized this as prompt instability, demonstrating that stochastic variations – such as the permutation of few-shot examples or minor syntactic shifts in instruction templates – can induce significant fluctuations in classification accuracy. This volatility suggests that the standard attention mechanism often converges on surface-level patterns rather than underlying logical structures. Furthermore, the architectural constraints of the transformer's context window present a dimensional bottleneck. As noted by Dong et al. \cite{dong_survey_2024}, fixed token limits necessitate a zero-sum trade-off between the depth of individual examples and the breadth of the reference set. In enterprise analytics, where datasets are high-dimensional and noisy, this limitation often leads to recency bias or the inclusion of non-representative outliers that confound the model's outcomes.

The SSAS framework departs from traditional ICL by replacing static, heuristically-derived prompts \cite{juros_llms_2024, siledar_one_2024} with a dynamically synthesized context. By applying a precision-filtering pipeline to the input background, we ensure that the hints provided to the model are mathematically optimized for representative signal. This transforms the context from a variable, human-engineered instruction into a stable, feature-engineered instrument, effectively addressing the inherent stochasticity of the generative process.

\subsection{Attention Mechanisms and the Signal-to-Noise Challenge}
The LLM Paradox identified in this study – wherein generative fluency inversely correlates with analytical precision – is fundamentally rooted in the transformer’s attention mechanism \cite{vaswani_attention_2023}. While the self-attention layer excels at global dependency modeling, its probabilistic nature becomes a liability when processing chaotic, non-curated datasets. In these environments, the model often fails to distinguish high-value signal from background noise, leading to a degradation of the latent space stability and consistency. Empirical evidence by Shi et al. \cite{shi_large_2023} suggests that the inclusion of irrelevant information within the input context is more detrimental to model performance than the omission of relevant data. This occurs because extraneous tokens force the attention mechanism to allocate significant weights to inconsequential patterns, effectively diluting the focus on salient features. This challenge aligns with the Information Bottleneck (IB) principle \cite{tishby_deep_2015}, which posits that an optimal learning system should maximize the compression of input data while retaining only the information most pertinent to the target output.

The SSAS methodology operationalizes the IB principle by implementing a pre-inference filtering stage involving data refinement. By calculating a Signal-to-Noise Ratio (SNR), the framework systematically suppresses irrelevant data and outliers before they are ingested by the LLM model. This intervention enforces a deterministic focus, ensuring that the transformer’s limited attention budget is reserved exclusively for contextually dense, representative data points. Consequently, the methodology bridges the gap between the stochasticity of generative architectures and the invariance required for enterprise-grade analytics.

\subsection{Hierarchical Information and Semantic Alignment}
The challenge of organizing large-scale, chaotic datasets into actionable taxonomies has historically been addressed through clustering and dimensionality reduction. However, traditional methods often treat summarization as a flat, single-pass task, which fails to capture the multi-layered nature of enterprise data. Standard Retrieval-Augmented Generation (RAG) and recursive summarization techniques often suffer from information dilution, where the specific nuances of data points are lost during mid-level aggregation \cite{wang_recursively_2025}. While recent work in Coarse-to-Fine Retrieval has improved document-level QA, these models frequently lack a mechanism to reconcile strategic top-down intent with bottom-up empirical evidence. Our SSAS framework addresses this by implementing a dual-flow logic that ensures narrative consistency across three distinct levels: Themes, Stories, and Clusters.

The integration of syntactic alignment (structural hierarchy) \cite{ma_structural_2016} and semantic alignment \cite{jurafsky_speech_2014, nan_semantic_2016} (latent meaning) is a recognized frontier in Natural Language Processing (NLP) \cite{maas_learning_2011, socher_recursive_2013, turney_thumbs_2001}. While Named Entity Recognition (NER) \cite{roy_recent_2021} and Topic Modeling \cite{jelodar_latent_2019, xun_topic_2016} provide semantic labels, they do not inherently weight the importance of data points based on their representative power within a broader narrative. Our approach draws inspiration from Selective Attention mechanisms in cognitive modeling \cite{desimone_neural_1995}, where non-essential noise is suppressed prior to high-level cognitive processing. By utilizing the Summary-of-Summaries (SoS) architecture, SSAS creates a bounded attention environment that forces the LLM to focus on distilled, sentiment-dense narratives \cite{sharma_movie_2023} rather than being distracted by the stochastic variance of raw, unweighted text.

Finally, our work builds upon the Information Bottleneck (IB) principle \cite{tishby_deep_2015}, which suggests that an optimal analytical model must compress input to retain only the features most relevant to the target output. While generative AI is optimized for creative novelty, the SSAS methodology enforces analytical integrity by treating the input context as a precision-engineered feature set. This transforms the LLM outputs, into consistent outcomes using SSAS precision-filtering, thus addressing the malady of unpredictability often cited in current enterprise AI research \cite{davenport_all-ai_2023}.

\section{Syntactic \& Semantic Context Assessment Summarization (SSAS)}
The strategic rationale behind our SSAS methodology \cite{mayande_syntactic_2024} is the implementation of a bounded attention mechanism. By pre-processing raw text through synchronized syntactic and semantic filters, we constrain the LLM’s focus to high-signal tokens, effectively performing feature engineering at the prompt level \cite{siledar_one_2024}. This ensures that the model recognizes the structural hierarchy of the data before the semantic layer interprets the underlying mood or sentiment following the Compositional Semantics principle \cite{gleitman_invitation_1995, sun_computational_2024}. Table \ref{tab:syntactic_semantic} contrasts the two foundational pillars of our methodology i.e. syntactic alignment and semantic alignment \cite{ma_structural_2016, jurafsky_speech_2014, nan_semantic_2016, holbrook_unified_1994, tan_scalable_2012}.

\begin{table*}[t]
\centering
\renewcommand{\arraystretch}{1.4} 
\begin{tabular}{>{\centering\arraybackslash}p{6.5cm} >{\centering\arraybackslash}p{6.5cm}}
\toprule
\textbf{Syntactic Alignment} & \textbf{Semantic Alignment} \\ \midrule
Defines structural rules governing how words combine into grammatical sentences. Acts as a mechanism for models to understand the structural hierarchy of data. & 
Delves into the meaning of words and sentences. Explores how syntactic structures map onto semantic roles to extract the ``who, what, when, where, and why'' of data. \\ \addlinespace

\textbf{Examples:} & \textbf{Examples:} \\
Coarse-to-Fine Retrieval, Spatial Reasoning Improvements \cite{ranasinghe_learning_2024}, Efficient Tuning for Document Visual QA \cite{lewis_generative_2019} & Word Embeddings \cite{mikolov_efficient_2013}, Named Entity Recognition (NER) \cite{roy_recent_2021}, Topic Modeling \cite{jelodar_latent_2019, xun_topic_2016} \\
\bottomrule
\end{tabular}
\vspace{8pt}
\caption{Syntactic vs. Semantic Alignment}
\label{tab:syntactic_semantic}
\end{table*}

By combining these alignments, our methodology creates an accurate, distilled summary of the dataset. This summary functions as a specific input prompt \cite{siledar_one_2024} that focuses the LLM's attention mechanism \cite{vaswani_attention_2023} to the provided essential information rather than being distracted by the surrounding noise \cite{shi_large_2023}. This alignment is optimized when applied across a rigorous data hierarchy. 

\subsection{Hierarchical Data Classification: Themes, Stories, and Clusters}
To transform large-scale, chaotic datasets into actionable insights, a structured hierarchical classification is necessary. Our framework ensures that every data point is evaluated for its contribution to the macro-narrative, preventing the loss of signal in high-volume environments. The SSAS methodology implements three distinct levels:
\begin{enumerate}
    \item Themes: The most general classification level, identifying the primary macro-topic across all data points within the set.
    \item Stories: The intermediate level of classification, ensuring narrative consistency by identifying specific sub-topics within a theme.
    \item Clusters: The lowest level of classification, where the algorithm utilizes localized precision to identify similar data points.
\end{enumerate}

\begin{figure}[t] 
    \centering
    \includegraphics[width=0.75\textwidth]{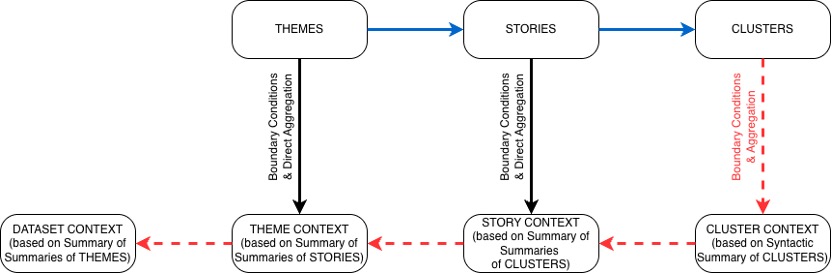} 
    \caption{SSAS Architecture for Context Assessment}
    \label{fig:ssas_architecture}
\end{figure}

The architecture operates on a dual-flow logic that reconciles strategic intent with empirical evidence, shown in Figure \ref{fig:ssas_architecture}. The first is the top-down taxonomy, i.e. strategic intent, which is the classification flow of Themes $\rightarrow$ Stories $\rightarrow$ Clusters, organizing the data into increasingly granular, manageable segments \cite{im_hierarchical_2023}. The second  is the bottom-up aggregation, i.e. data evidence, which gives the insight flow of Cluster Contexts $\rightarrow$ Story Contexts $\rightarrow$ Theme Contexts, aggregating data to build the Summary of Summaries (SoS). This ensuring that high-level insights are grounded in the localized precision of the underlying clusters \cite{christensen_hierarchical_2014}.

\subsection{Summary-of-Summaries (SoS)}
The implementation culminates in the context-localized Summary-of-Summaries (SoS) architecture. This approach on data pre-processing strategically bounds the LLM's focus by providing a concise, iterative summary as the primary input prompt \cite{siledar_one_2024}, effectively reducing the probability of stochastic drift – a phenomenon where the model loses its objective over long context windows \cite{wang_recursively_2025}. The process follows a specific aggregation path:
\begin{enumerate}
    \item Cluster Context: A syntactic summary of individual clusters
    \item Story Context: An aggregated summary based on the summary of cluster-summaries within that story \cite{christensen_hierarchical_2014}
    \item Theme Context: An aggregated summary based on the summary of story-summaries within that theme \cite{christensen_hierarchical_2014}
\end{enumerate}

The strategic value of SoS is its role as a feature engineering step. By distilling raw text into a sentiment-dense narrative, we force the LLM to align with core structures and factual content mitigating the risk of distraction from irrelevant tokens \cite{li_compressing_2023}.

\subsection{Signal-to-Noise Ratio (SNR)}
Maintaining data integrity requires a rigorous weighting algorithm to isolate high-value signal from noise. SSAS uses a weighting logic to validate data points across the hierarchical strata. The primary metric is the Signal to Noise Ratio ($SNR_i$), which is the weighted aggregate of three distinct dimensions:
\begin{equation}
\label{eq:snr}
SNR_i = \sum (S_{\text{Cluster}} + S_{\text{Story}} + S_{\text{Theme}})
\end{equation}
where:
\begin{itemize}
    \item $S_{\text{Cluster}}$ -- Cluster Signal to Noise Ratio that measures localized precision e.g. does it fit the immediate group?
    \item $S_{\text{Story}}$ -- Story Signal to Noise Ratio that measures narrative consistency e.g. does it fit the sub-topic?
    \item $S_{\text{Theme}}$ -- Theme Signal to Noise Ratio that measures global alignment e.g. does it fit the macro-topic?
\end{itemize}

Further, the methodology incorporates Weighted Amplitude, where keywords are weighted by frequency to enhance the signal. The outcome is Precision Filtering, which suppresses data points that share keywords but lack the contextual depth required for stable attention \cite{shi_large_2023}. This ensures the LLM is prompted only with high-signal data that fits the hierarchy perfectly, preventing out-of-context noise.

\subsection{Noise Mitigation: Irrelevant Data, Outlier Management}
Reliable sentiment prediction \cite{maas_learning_2011, niu_review_2021} necessitates aggressive noise removal to prevent the dilution of the model's focus. SSAS categorizes noise into ranked ordering in two steps:
\begin{enumerate}
    \item Irrelevant Data: This comprises data that does not fit into any defined classification level. Our algorithm labels these as irrelevant data.
    \item Outlier Data: These are data points within the classification levels that have a negligible impact on the whole level.
\end{enumerate}

Through this rank-ordering, the most representative and contextually dense data points are elevated, while outliers and irrelevant data are suppressed to the bottom of the dataset. This cleansing process ensures the final summary is laden with sentiment-rich, factual content, mitigating the risk of inconsistent outputs.

Based on the methodology details provided, we propose three hypotheses to explore:

\textit{Hypothesis 1:}
Datasets, in large-scale text analytics, are normally chaotic and noisy with inconsequential information. Classifying datasets into a Theme-Story-Cluster hierarchy can help make sense of large and chaotic data. It also helps identify the dataset parts that are signal versus noise. Identifying the noise helps focus the analysis on the signal within the dataset.

\textit{Hypothesis 2: }
LLMs struggle to produce consistent sentiment predictions. Adding a concise summary, as input, can enhance alignment with the data. It ensures LLMs focus on the core structure and factual content, while the summary itself highlights sentiment-laden words or phrases. This combined approach improves an LLM’s ability to interpret sentiment of original text.

\textit{Hypothesis 3: }
The SSAS methodology enables data classification, noise reduction and better sentiment prediction through the use of concise summaries as input. All these together enable better quality and more consistent results.

\section{Evaluation Framework}
\subsection{Frameworks}
To evaluate our hypotheses, we compare our SSAS methodology against a direct-LLM (DIRECT) approach. The DIRECT method defines the control group where raw datasets are fed directly into the LLM with a standard sentiment prediction prompt without any structural refinement. In comparison, in the SSAS method, data is systematically classified into a hierarchical structure of Themes, Stories, and Clusters before being presented to the LLM. The model receives both the specific data point and a classification summary, providing a framework for consistent reasoning.

We executed a total of 10 runs for each framework and compared the sentiment prediction given for each data point within a dataset. For this evaluation, we selected the Gemini 2.0 Flash Lite \cite{team_gemini_2025} as our primary LLM model. We selected this model because it represents the optimal balance between latency and reasoning capability, making it a high-pressure test case for consistency in high-efficiency, production-grade environments where output volatility is most likely to occur.

\subsection{Datasets}
To evaluate both frameworks, we chose three industry-standard datasets, compiled by the University of California, San Diego. These datasets were selected to test different semantic environments. Table \ref{tab:datasets_characterization} shows the datasets and their relevant information.

\begin{table}[t]
\centering
\small 
\renewcommand{\arraystretch}{1.5}
\setlength{\tabcolsep}{4pt} 

\begin{tabularx}{\textwidth}{>{\centering\arraybackslash\hsize=0.7\hsize}X c c >{\centering\arraybackslash\hsize=1.3\hsize}X}
\toprule
\textbf{Dataset Name \& Category} & \textbf{Scale} & \textbf{Temporal Range} & \textbf{Strategic Intent} \\ \midrule

\textbf{Amazon Product Reviews} \newline (Health \& Personal Care) & 
\makecell{155,745 reviews \\ 16,395 Stores} & 
01/01/2020 -- 05/23/2023 & 
Product-related, high-turnover sentiment analysis. \\ \addlinespace

\textbf{Google Business Reviews} \newline (American \& Fast Food) & 
\makecell{121,826 reviews \\ 251 Businesses} & 
03/01/2009 -- 08/25/2021 & 
Restaurant-related, service-heavy performance data. \\ \addlinespace

\textbf{Goodreads Book Reviews} \newline (General Books) & 
\makecell{157,407 reviews \\ 19,725 Book Titles} & 
12/07/2006 -- 11/03/2017 & 
Literary, subjective, and nuanced content for sentiment testing. \\ 
\bottomrule
\end{tabularx}
\vspace{8pt}
\caption{Datasets Characterization}
\label{tab:datasets_characterization}
\end{table}

\subsection{Dataset Characterization}
Framework robustness needs to be validated across diverse sector-specific behaviors. The volume distribution and temporal range of data significantly impact an LLM's ability to maintain a stable analytical phase. To ensure a consistent longitudinal analysis, data for all three sources were aggregated on a quarterly basis. The analysis focuses on two primary dimensions: Normalized Volume and Review Distribution, facilitating a standardized comparison of engagement patterns across different domains. The datasets vary significantly in their temporal depth. Table \ref{tab:quarterly_distribution} summarizes the longitudinal coverage for each dataset. Data Integrity Note: It should be noted that the initial and terminal quarters for each dataset may not represent full three-month periods. These variances are inherent to the data collection timestamps and the specific start/end dates of the scraping or archival processes.

\begin{table}[t]
\centering
\renewcommand{\arraystretch}{1.4}
\begin{tabular}{l c c} 
\toprule
\textbf{Dataset} & \textbf{Primary Entity} & \textbf{Total Quarters Analyzed} \\ \midrule
Amazon Product Reviews  & Stores      & 14 \\ \addlinespace
Goodreads Book Reviews  & Book Titles & 45 \\ \addlinespace
Google Business Reviews & Restaurants & 51 \\ \bottomrule
\end{tabular}
\vspace{8pt}
\caption{Dataset Quarterly Distribution}
\label{tab:quarterly_distribution}
\end{table}

To further characterize the datasets, we looked at the quarterly activity of the primary entities (Stores, Book Titles, and Restaurants) across two distinct metrics:
\begin{enumerate}
    \item Normalized Volume: This metric measures the intensity of review activity relative to the dataset average. For each entity, the volume was normalized and then binned into binary categories:
    \begin{itemize}
        \item HIGH: Data points exceeding the calculated mean of the Normalized Volume.
        \item LOW: Data points at or below the calculated mean.
    \end{itemize}
    \item Review Distribution: Review Distribution assesses the continuity of engagement by calculating the frequency of active quarters (defined as any quarter containing at least one review). Entities were then segmented into three density buckets:
    \begin{itemize}
        \item 100\% (Persistent): Reviews present in every available quarter of the life-cycle of the data set
        \item 51\% to 99\% (Intermittent): Reviews present in more than half, but not all, quarters
        \item 0\%–50\% (Sparse): Reviews present in half or fewer quarters
    \end{itemize}
\end{enumerate}

\subsection{The "Noise vs. Signal" Identification Process}
A core component of SSAS is the identification of outliers using a cluster-level Cumulative SNR (defined below). To identify and exclude noise, the following process is applied:
\begin{enumerate}
    \item Normalized Volume: The count of cluster data points is normalized on a scale of 0 to 100.
    \item Cumulative SNR: A cumulative Signal to Noise Ratio (SNR) is generated, for each cluster, by sum of individual SNR of all data points within that cluster.
    \item SNR Normalization: The cluster SNR values are normalized on a scale of 0 to 100.
    \item Gate Threshold: Clusters are evaluated against a threshold of 0.1. Only clusters where either the normalized Volume or the normalized SNR is 0.1 or higher are retained.
    \item Outlier Identification: Data points within clusters falling below this threshold are flagged as outliers and excluded to ensure semantic integrity.
\end{enumerate}

\subsection{Comparative Analysis of Robustness Scenarios}
To stress-test the methodology beyond the Base (all data) case, we established 6 Robustness Scenarios. These represent a logical progression from optimal data environments to extreme "Low Volume/Low Distribution" edge cases. The 6 Robustness Scenarios are (1) High Volume with 100\% Review Distribution, (2) High Volume with (99–51)\% Review Distribution, (3) High Volume with (0–50)\% Review Distribution, (4) Low Volume with 100\% Review Distribution, (5) Low Volume with (99–51)\% Review Distribution, and (6) Low Volume  – (0–50)\% Review Distribution.

The overview of the Design of Experiments is shown in Table \ref{tab:design_of_experiments}.

\begin{table*}[t!]
\centering
\begin{minipage}{0.95\textwidth} 
    \scriptsize
    \renewcommand{\arraystretch}{1.6} 

    \begin{tabularx}{\textwidth}{
        >{\centering\arraybackslash}p{1.8cm} 
        >{\centering\arraybackslash}p{1cm} 
        >{\centering\arraybackslash}p{1.2cm} 
        >{\centering\arraybackslash}p{1.5cm} 
        >{\centering\arraybackslash}X 
        >{\centering\arraybackslash}X 
        >{\centering\arraybackslash}X 
        >{\centering\arraybackslash}p{3.2cm}
    }
    \toprule
    \multirow{2}{*}{\textbf{Scenario}} & \multicolumn{2}{c}{\textbf{Input}} & \multirow{2}{*}{\textbf{Dataset}} & \multicolumn{3}{c}{\textbf{Output}} & \multirow{2}{*}{\textbf{Design Rationale}} \\
    \cmidrule(lr){2-3} \cmidrule(lr){5-7}
     & \textbf{Volume} & \textbf{Dist.} & & \textbf{ALL} & \textbf{w/o Irr.} & \textbf{w/o I+O} & \\ 
    \midrule

    \textbf{\makecell[c]{Base (Themes)}} & \textbf{ALL} & \textbf{ALL} & \makecell[c]{Amazon\\Google\\Goodreads} & \multicolumn{3}{c}{\makecell[c]{Divided into $N$ Themes; reporting on\\\#Stories, \#Clusters, and \#Datapoints}} & See Comment 1, 4 \\
    \midrule

    \textbf{Base} & \textbf{ALL} & \textbf{ALL} & \makecell[c]{Amazon\\Google\\Goodreads} & 
    \multirow{7}{\linewidth}{\centering Appropriate samples included} & 
    \multirow{7}{\linewidth}{\centering Appropriate sub-samples included} & 
    \multirow{7}{\linewidth}{\centering Appropriate sub-samples included} & See Comments 1, 4--7 \\
    \cmidrule(r){1-4} \cmidrule(l){8-8}

    \textbf{Scenario 1} & \multirow{3}{*}{\textbf{HIGH}} & \textbf{100\%} & \makecell[c]{Amazon\\Google\\Goodreads} & & & & \multirow{9}{3.2cm}{\centering See Comments 1, 2, 4--7 and sub-notes} \\
    \cmidrule(r){1-1} \cmidrule(r){3-4}
    \textbf{Scenario 2} & & \textbf{51--99\%} & \makecell[c]{Amazon\\Google\\Goodreads} & & & & \\
    \cmidrule(r){1-1} \cmidrule(r){3-4}
    \textbf{Scenario 3} & & \textbf{0--50\%} & \makecell[c]{Amazon\\Google\\Goodreads} & & & & \\
    \cmidrule(r){1-4} \cmidrule(l){8-8}

    \textbf{Scenario 4} & \multirow{3}{*}{\textbf{LOW}} & \textbf{100\%} & \makecell[c]{Amazon\\Google\\Goodreads} & & & & \multirow{9}{3.2cm}{\centering See Comments 1, 3, 4--7 and sub-notes} \\
    \cmidrule(r){1-1} \cmidrule(r){3-4}
    \textbf{Scenario 5} & & \textbf{51--99\%} & \makecell[c]{Amazon\\Google\\Goodreads} & & & & \\
    \cmidrule(r){1-1} \cmidrule(r){3-4}
    \textbf{Scenario 6} & & \textbf{0--50\%} & \makecell[c]{Amazon\\Google\\Goodreads} & & & & \\
    \bottomrule
    \end{tabularx}

    \vspace{10pt}

    {\scriptsize \raggedright
    \textbf{NOTES:}\\
    This is a full factorial experiment such that all possible data input and output scenarios are considered as shown in the table above.
    
    \vspace{4pt}
    \textbf{Variation in Input Parameters:}\\
    \textbf{Comment 1:} Allows comparison of 3 different, industry-standard datasets with varying contexts.\\
    \textbf{Comment 2:} Robustness checks \textit{vis-\`a-vis} the Base case for \textbf{HIGH} volume:\\
    \hspace*{1em} 1. HIGH segments data into High volume to compare against the Base case.\\
    \hspace*{1em} 2. Review Distribution disaggregates HIGH into 100\%, 51--99\%, or 0--50\% frequency.\\
    \textbf{Comment 3:} Robustness checks \textit{vis-\`a-vis} the Base case for \textbf{LOW} volume:\\
    \hspace*{1em} 1. LOW segments data into Low volume to compare against the Base case.\\
    \hspace*{1em} 2. Review Distribution disaggregates LOW into 100\%, 51--99\%, or 0--50\% frequency.
    
    \vspace{4pt}
    \textbf{Variation in Output Parameters:}\\
    \textbf{Comment 4:} Outputs include Net Consistency, Data Conditioning, and Sub-Total at three stages: All data, w/o Irrelevant, and w/o Irrelevant+Outlier.
    
    \vspace{4pt}
    \textbf{Definition of Consistency Metrics:}\\
    \textbf{Comment 5:} \textit{Net Consistency}: improvements seen for SSAS (vs. DIRECT) across 10 independent runs.\\
    \textbf{Comment 6:} \textit{Data Conditioning}: improvements achieved through the removal of noise (Irrelevant/Outlier data).\\
    \textbf{Comment 7:} \textit{Sub-Total}: the cumulative sum of improvements from both Net Consistency and Data Conditioning. \par}

    \vspace{12pt} 
    \caption{Design of Experiments}
    \label{tab:design_of_experiments}
\end{minipage}
\end{table*}

\section{Results}
\subsection{Hierarchical Data Classification: Themes, Stories, and Clusters}
Table \ref{tab:dataset_classification} shows the hierarchical classification of the datasets into Themes, Stories, and Clusters. The table also provides a breakdown of the number of stories and clusters within each theme, along with the total data points assigned to those categories. The aggregate totals for each dataset are highlighted in RED at the conclusion of each dataset section to provide a comprehensive summary of the methodology’s output.

\begin{table*}[t!]
\centering
\scriptsize 
\renewcommand{\arraystretch}{1.2}
\setlength{\tabcolsep}{1.5pt} 

\begin{tabularx}{\textwidth}{@{} >{\raggedright\arraybackslash}p{6em} *{12}{>{\centering\arraybackslash}X} @{}}
\toprule
 & \multicolumn{4}{c}{\textbf{All Data}} & \multicolumn{4}{c}{\textbf{w/o Irrelevant}} & \multicolumn{4}{c}{\textbf{w/o Irrelevant, Outlier}} \\
\cmidrule(lr){2-5} \cmidrule(lr){6-9} \cmidrule(lr){10-13}
\textbf{Dataset} & \textbf{Thm.} & \textbf{\#St.} & \textbf{\#Cl.} & \textbf{\#Pts.} & \textbf{Thm.} & \textbf{\#St.} & \textbf{\#Cl.} & \textbf{\#Pts.} & \textbf{Thm.} & \textbf{\#St.} & \textbf{\#Cl.} & \textbf{\#Pts.} \\ 
\midrule

\rowcolor{white} \cellcolor{white} & -1 & 1 & 344 & 637 & - & - & - & - & - & - & - & - \\
\cellcolor{white} & 0 & 11 & 6290 & 46974 & 0 & 10 & 6290 & 44945 & 0 & 10 & 649 & 35384 \\
\rowcolor{white} \cellcolor{white} & 1 & 10 & 4814 & 41568 & 1 & 9 & 4813 & 39507 & 1 & 9 & 404 & 32664 \\
\cellcolor{white} & 2 & 10 & 2339 & 15464 & 2 & 9 & 2339 & 15095 & 2 & 9 & 243 & 11456 \\
\rowcolor{white} \cellcolor{white} & 3 & 10 & 1615 & 14701 & 3 & 9 & 1615 & 14511 & 3 & 9 & 195 & 12067 \\
\cellcolor{white} & 4 & 10 & 2273 & 14298 & 4 & 9 & 2273 & 14033 & 4 & 9 & 204 & 10445 \\
\rowcolor{white} \cellcolor{white} & 5 & 10 & 1602 & 10803 & 5 & 9 & 1602 & 10696 & 5 & 9 & 160 & 8234 \\
\cellcolor{white} \multirow{-8}{*}{\makecell[l]{\textbf{Amazon}\\\textbf{Product}\\\textbf{Reviews}}} & 6 & 10 & 1249 & 4204 & 6 & 9 & 1249 & 4138 & 6 & 8 & 63 & 2242 \\
\rowcolor{white} \cellcolor{white} & 7 & 10 & 724 & 2351 & 7 & 9 & 724 & 2163 & 7 & 8 & 34 & 1215 \\
\cellcolor{white} & 8 & 4 & 668 & 2251 & 8 & 3 & 668 & 2249 & 8 & 3 & 46 & 1200 \\
\rowcolor{white} \cellcolor{white} & 9 & 3 & 265 & 803 & 9 & 2 & 265 & 795 & 9 & 2 & 21 & 419 \\
\cellcolor{white} & 10 & 2 & 216 & 515 & 10 & 2 & 216 & 515 & 10 & 1 & 5 & 195 \\
\rowcolor{white} \cellcolor{white} & 11 & 6 & 215 & 486 & 11 & 6 & 215 & 486 & 11 & 3 & 3 & 148 \\
\cellcolor{white} & 12 & 4 & 68 & 396 & 12 & 4 & 68 & 396 & 12 & 4 & 5 & 306 \\
\rowcolor{white} \cellcolor{white} & 13 & 2 & 109 & 294 & 13 & 2 & 109 & 294 & 13 & 2 & 2 & 127 \\
\rowcolor{white} \cellcolor{white} Total & {\color{red}15} & {\color{red}103} & {\color{red}22791} & {\color{red}155745} & {\color{red}14} & {\color{red}92} & {\color{red}22446} & {\color{red}149823} & {\color{red}14} & {\color{red}86} & {\color{red}2034} & {\color{red}116102} \\
\midrule

\rowcolor{white} \cellcolor{white} & -1 & 1 & 147 & 273 & - & - & - & - & - & - & - & - \\
\cellcolor{white} & 0 & 11 & 3978 & 88107 & 0 & 10 & 3978 & 84789 & 0 & 10 & 190 & 74451 \\
\rowcolor{white} \cellcolor{white} & 1 & 10 & 1146 & 10668 & 1 & 9 & 1146 & 10400 & 1 & 5 & 40 & 7631 \\
\cellcolor{white} & 2 & 10 & 919 & 9299 & 2 & 9 & 919 & 9059 & 2 & 7 & 27 & 7066 \\
\rowcolor{white} \cellcolor{white} & 3 & 10 & 564 & 3257 & 3 & 9 & 564 & 3184 & 3 & 6 & 9 & 1917 \\
\cellcolor{white} & 4 & 10 & 435 & 2506 & 4 & 9 & 435 & 2471 & 4 & 4 & 14 & 1386 \\
\rowcolor{white} \cellcolor{white} & 5 & 10 & 286 & 2145 & 5 & 9 & 286 & 2113 & 5 & 7 & 10 & 1563 \\
\cellcolor{white} \multirow{-8}{*}{\makecell[l]{\textbf{Google}\\\textbf{Business}\\\textbf{Reviews}}} & 6 & 10 & 284 & 1556 & 6 & 9 & 284 & 1531 & 6 & 4 & 5 & 950 \\
\rowcolor{white} \cellcolor{white} & 7 & 9 & 311 & 992 & 7 & 8 & 311 & 965 & 7 & 2 & 3 & 230 \\
\cellcolor{white} & 8 & 2 & 182 & 842 & 8 & 2 & 182 & 842 & 8 & 1 & 3 & 444 \\
\rowcolor{white} \cellcolor{white} & 9 & 11 & 196 & 804 & 9 & 10 & 196 & 803 & 9 & 4 & 4 & 200 \\
\cellcolor{white} & 10 & 5 & 70 & 650 & 10 & 5 & 70 & 650 & 10 & 2 & 3 & 486 \\
\rowcolor{white} \cellcolor{white} & 11 & 5 & 129 & 370 & 11 & 5 & 129 & 370 & 11 & 2 & 2 & 110 \\
\cellcolor{white} & 12 & 5 & 105 & 198 & 12 & 4 & 105 & 196 & - & - & - & - \\
\rowcolor{white} \cellcolor{white} & 13 & 4 & 52 & 159 & 13 & 3 & 52 & 155 & - & - & - & - \\
\rowcolor{white} \cellcolor{white} Total & {\color{red}15} & {\color{red}113} & {\color{red}8804} & {\color{red}121826} & {\color{red}14} & {\color{red}101} & {\color{red}8657} & {\color{red}117528} & {\color{red}12} & {\color{red}54} & {\color{red}310} & {\color{red}96434} \\
\midrule

\rowcolor{white} \cellcolor{white} & -1 & 1 & 3701 & 12443 & - & - & - & - & - & - & - & - \\
\cellcolor{white} & 0 & 11 & 11182 & 116359 & 0 & 10 & 11182 & 115048 & 0 & 10 & 1349 & 99216 \\
\rowcolor{white} \cellcolor{white} & 1 & 10 & 1143 & 4733 & 1 & 9 & 1143 & 4691 & 1 & 6 & 78 & 3081 \\
\cellcolor{white} & 2 & 3 & 218 & 605 & 2 & 3 & 218 & 605 & 2 & 3 & 14 & 272 \\
\rowcolor{white} \cellcolor{white} & 3 & 10 & 1480 & 6272 & 3 & 9 & 1480 & 6167 & 3 & 9 & 117 & 4106 \\
\cellcolor{white} \multirow{-6}{*}{\makecell[l]{\textbf{Goodreads}\\\textbf{Book}\\\textbf{Reviews}}} & 4 & 10 & 905 & 2335 & 4 & 9 & 905 & 2263 & 4 & 9 & 43 & 1031 \\
\rowcolor{white} \cellcolor{white} & 5 & 2 & 265 & 714 & 5 & 2 & 265 & 714 & 5 & 2 & 11 & 356 \\
\cellcolor{white} & 6 & 7 & 760 & 4096 & 6 & 6 & 760 & 4095 & 6 & 4 & 55 & 3033 \\
\rowcolor{white} \cellcolor{white} & 7 & 5 & 451 & 1442 & 7 & 5 & 451 & 1442 & 7 & 5 & 30 & 832 \\
\cellcolor{white} & 8 & 7 & 715 & 2296 & 8 & 6 & 715 & 2294 & 8 & 6 & 42 & 1262 \\
\rowcolor{white} \cellcolor{white} & 9 & 10 & 775 & 2401 & 9 & 9 & 775 & 2370 & 9 & 9 & 46 & 1369 \\
\cellcolor{white} & 10 & 4 & 791 & 3711 & 10 & 4 & 791 & 3711 & 10 & 3 & 57 & 2575 \\
\rowcolor{white} \cellcolor{white} Total & {\color{red}12} & {\color{red}80} & {\color{red}22386} & {\color{red}157407} & {\color{red}11} & {\color{red}72} & {\color{red}18685} & {\color{red}143400} & {\color{red}11} & {\color{red}66} & {\color{red}1842} & {\color{red}117133} \\
\bottomrule
\end{tabularx}

\vspace{10pt} 
\caption{Dataset Classification into Themes, Stories and Clusters}
\label{tab:dataset_classification}

\end{table*}

\subsubsection{Noise Classification: Irrelevant and Outlier data}
Within Table \ref{tab:dataset_classification}, we also list and remove the two types of noise we initially defined:
\begin{itemize}
    \item Irrelevant Data: Data points identified as irrelevant are grouped under a specialized labeled "-1" within all three hierarchical levels. These data points are shown in the ‘All Data’ section but removed for the other sub-sections.
    \item Outlier Data: Individual data points that do not conform to the specific requirements for inclusion within a Story or Cluster (details in Appendix A) are removed in the ‘w/o Irrelevant, Outlier’ section of the table.
\end{itemize}

As seen from the aggregates at the end of each dataset, the number of Themes, Stories and Clusters (along with the associated data points) reduce from ‘All Data’ to ‘w/o Irrelevant’ to  ‘w/o Irrelevant, Outlier’ sections for the same dataset. Most times, we see a significant reduction in the number of stories and clusters when moving away from ‘All Data’. In some cases, like Google Business Reviews, we can also see a significant effect on themes. In this dataset, the outlier detection phase of our methodology identified that all data points within Themes 12 and 13 were outliers. Consequently, these themes were removed in the final analysis, reducing the total theme count to 12 from the original 15.

This tiered approach allows for a transparent view of how our methodology isolates meaningful content from noise, ensuring that the final clusters and stories are derived from the most pertinent data points.

\subsection{Consistency Improvements: SSAS vs. DIRECT}
Table \ref{tab:consistency_improvements} presents the performance results, of SSAS vs. DIRECT methodology, across the Base plus 6 Robustness Scenarios. The evaluation is categorized by the Input Controls and Datasets, with outputs tracked across the three stages of refinement: All Data, ‘w/o Irrelevant’ Data, and ‘w/o Irrelevant, Outlier’ Data. The core of our evaluation rests in the ‘Consistency Improvements in Output Data points (vs. DIRECT)’ section, which is divided into three critical components:
\begin{itemize}
    \item Net Consistency: Measures the consistency of sentiment prediction, SSAS vs. DIRECT, across 10 independent runs.
    \item Data Conditioning: Quantifies gains achieved through the removal of noisy (Irrelevant and Outlier) data points.
    \item Total Improvement: The cumulative improvement derived from both metrics above.
\end{itemize}

\begin{table*}[t!]
\centering
\scriptsize
\renewcommand{\arraystretch}{1.3}
\setlength{\tabcolsep}{1.2pt} 

\definecolor{condRed}{HTML}{8C3626}   
\definecolor{condGreen}{HTML}{385623} 
\definecolor{condBlue}{HTML}{1F3864}  

\begin{tabularx}{\textwidth}{@{} >{\centering\arraybackslash}p{4em} >{\centering\arraybackslash}p{3.5em} >{\centering\arraybackslash}p{3.5em} >{\centering\arraybackslash}p{4em} *{3}{>{\centering\arraybackslash}X} *{9}{>{\centering\arraybackslash}X} >{\centering\arraybackslash}p{5.5em} @{}}
\toprule
\multirow{3}{*}{\textbf{Scenario}} & \multicolumn{2}{c}{\textbf{Input Control}} & \multirow{3}{*}{\textbf{Dataset}} & \multicolumn{3}{c}{\textbf{Output Datapoints}} & \multicolumn{9}{c}{\textbf{Consistency Improvements in Output Datapoints (vs. DIRECT)}} & \multirow{3}{*}{\makecell{\textbf{Supporting}\\\textbf{Evidence}}} \\
\cmidrule(lr){2-3} \cmidrule(lr){5-7} \cmidrule(lr){8-16}
 & \multirow{2}{*}{\textbf{Volume}} & \multirow{2}{*}{\makecell{\textbf{Review}\\\textbf{Dist.}}} & & \multirow{2}{*}{\textbf{ALL}} & \multirow{2}{*}{\makecell{\textbf{w/o}\\\textbf{Irrel.}}} & \multirow{2}{*}{\makecell{\textbf{w/o}\\\textbf{Ir+Out}}} & \multicolumn{3}{c}{\textbf{ALL}} & \multicolumn{3}{c}{\textbf{w/o Irrelevant}} & \multicolumn{3}{c}{\textbf{w/o Irrel, Outlier}} & \\
\cmidrule(lr){8-10} \cmidrule(lr){11-13} \cmidrule(lr){14-16}
 & & & & & & & \textbf{\tiny Net Cons.} & \textbf{\tiny Data Cond.} & \textbf{\tiny Total Improv.} & \textbf{\tiny Net Cons.} & \textbf{\tiny Data Cond.} & \textbf{\tiny Total Improv.} & \textbf{\tiny Net Cons.} & \textbf{\tiny Data Cond.} & \textbf{\tiny Total Improv.} & \\
\midrule

\multirow{3}{*}{\textbf{Base}} & \multirow{3}{*}{\textbf{ALL}} & \multirow{3}{*}{\textbf{ALL}} 
 & \textbf{Amazon} & 155745 & 149823 & 116102 & \color{condRed}3.6\% & \color{condGreen}0.0\% & \color{condBlue}3.6\% & \color{condRed}3.5\% & \color{condGreen}3.8\% & \color{condBlue}7.3\% & \color{condRed}2.5\% & \color{condGreen}25.5\% & \color{condBlue}28.0\% & Appendix A.1.1 \\
 & & & \textbf{Google} & 121826 & 117528 & 96434 & \color{condRed}2.5\% & \color{condGreen}0.0\% & \color{condBlue}2.5\% & \color{condRed}2.5\% & \color{condGreen}3.5\% & \color{condBlue}6.0\% & \color{condRed}1.1\% & \color{condGreen}20.8\% & \color{condBlue}22.0\% & Appendix A.1.2 \\
 & & & \textbf{Goodreads} & 157407 & 143400 & 117133 & \color{condRed}1.8\% & \color{condGreen}0.0\% & \color{condBlue}1.8\% & \color{condRed}1.2\% & \color{condGreen}8.9\% & \color{condBlue}10.1\% & \color{condRed}1.1\% & \color{condGreen}25.6\% & \color{condBlue}26.7\% & Appendix A.1.3 \\
\midrule

\multirow{3}{*}{\textbf{Scenario 1}} & \multirow{3}{*}{\textbf{HIGH}} & \multirow{3}{*}{\textbf{100}} 
 & \textbf{Amazon} & 17131 & 16442 & 12646 & \color{condRed}4.1\% & \color{condGreen}0.0\% & \color{condBlue}4.1\% & \color{condRed}3.9\% & \color{condGreen}4.0\% & \color{condBlue}8.0\% & \color{condRed}3.0\% & \color{condGreen}26.2\% & \color{condBlue}29.2\% & Appendix A.2.1 \\
 & & & \textbf{Google} & 0 & 0 & 0 & \color{condRed}-- & \color{condGreen}-- & \color{condBlue}-- & \color{condRed}-- & \color{condGreen}-- & \color{condBlue}-- & \color{condRed}-- & \color{condGreen}-- & \color{condBlue}-- & \multirow{2}{*}{\makecell[c]{No Datapoints}} \\
 & & & \textbf{Goodreads} & 0 & 0 & 0 & \color{condRed}-- & \color{condGreen}-- & \color{condBlue}-- & \color{condRed}-- & \color{condGreen}-- & \color{condBlue}-- & \color{condRed}-- & \color{condGreen}-- & \color{condBlue}-- & \\
\midrule

\multirow{3}{*}{\textbf{Scenario 2}} & \multirow{3}{*}{\textbf{HIGH}} & \multirow{3}{*}{\textbf{51--99}} 
 & \textbf{Amazon} & 59858 & 57404 & 44441 & \color{condRed}3.8\% & \color{condGreen}0.0\% & \color{condBlue}3.8\% & \color{condRed}3.76\% & \color{condGreen}4.10\% & \color{condBlue}7.9\% & \color{condRed}3.1\% & \color{condGreen}25.8\% & \color{condBlue}28.8\% & Appendix A.3.1 \\
 & & & \textbf{Google} & 93640 & 90378 & 73612 & \color{condRed}2.8\% & \color{condGreen}0.0\% & \color{condBlue}2.8\% & \color{condRed}2.8\% & \color{condGreen}3.5\% & \color{condBlue}6.3\% & \color{condRed}1.4\% & \color{condGreen}21.4\% & \color{condBlue}22.8\% & Appendix A.3.2 \\
 & & & \textbf{Goodreads} & 18720 & 17332 & 14761 & \color{condRed}1.8\% & \color{condGreen}0.0\% & \color{condBlue}1.8\% & \color{condRed}1.2\% & \color{condGreen}7.4\% & \color{condBlue}8.6\% & \color{condRed}1.0\% & \color{condGreen}21.1\% & \color{condBlue}22.2\% & Appendix A.3.3 \\
\midrule

\multirow{3}{*}{\textbf{Scenario 3}} & \multirow{3}{*}{\textbf{HIGH}} & \multirow{3}{*}{\textbf{0--50}} 
 & \textbf{Amazon} & 42477 & 41203 & 32976 & \color{condRed}3.3\% & \color{condGreen}0.0\% & \color{condBlue}3.3\% & \color{condRed}3.2\% & \color{condGreen}3.0\% & \color{condBlue}6.2\% & \color{condRed}2.3\% & \color{condGreen}22.4\% & \color{condBlue}24.7\% & Appendix A.4.1 \\
 & & & \textbf{Google} & 5518 & 5368 & 4619 & \color{condRed}2.5\% & \color{condGreen}0.0\% & \color{condBlue}2.5\% & \color{condRed}2.4\% & \color{condGreen}2.7\% & \color{condBlue}5.1\% & \color{condRed}1.3\% & \color{condGreen}16.3\% & \color{condBlue}17.6\% & Appendix A.4.2 \\
 & & & \textbf{Goodreads} & 93213 & 84762 & 69706 & \color{condRed}1.9\% & \color{condGreen}0.0\% & \color{condBlue}1.9\% & \color{condRed}1.3\% & \color{condGreen}9.1\% & \color{condBlue}10.3\% & \color{condRed}1.2\% & \color{condGreen}25.2\% & \color{condBlue}26.4\% & Appendix A.4.3 \\
\midrule

\multirow{3}{*}{\textbf{Scenario 4}} & \multirow{3}{*}{\textbf{LOW}} & \multirow{3}{*}{\textbf{100}} 
 & \textbf{Amazon} & 0 & 0 & 0 & \color{condRed}-- & \color{condGreen}-- & \color{condBlue}-- & \color{condRed}-- & \color{condGreen}-- & \color{condBlue}-- & \color{condRed}-- & \color{condGreen}-- & \color{condBlue}-- & \multirow{3}{*}{\makecell[c]{No Datapoints}} \\
 & & & \textbf{Google} & 0 & 0 & 0 & \color{condRed}-- & \color{condGreen}-- & \color{condBlue}-- & \color{condRed}-- & \color{condGreen}-- & \color{condBlue}-- & \color{condRed}-- & \color{condGreen}-- & \color{condBlue}-- & \\
 & & & \textbf{Goodreads} & 0 & 0 & 0 & \color{condRed}-- & \color{condGreen}-- & \color{condBlue}-- & \color{condRed}-- & \color{condGreen}-- & \color{condBlue}-- & \color{condRed}-- & \color{condGreen}-- & \color{condBlue}-- & \\
\midrule

\multirow{3}{*}{\textbf{Scenario 5}} & \multirow{3}{*}{\textbf{LOW}} & \multirow{3}{*}{\textbf{51--99}} 
 & \textbf{Amazon} & 44 & 41 & 31 & \color{condRed}4.5\% & \color{condGreen}0.0\% & \color{condBlue}4.5\% & \color{condRed}0.0\% & \color{condGreen}6.8\% & \color{condBlue}6.8\% & \color{condRed}6.5\% & \color{condGreen}29.5\% & \color{condBlue}36.0\% & Appendix A.5.1 \\
 & & & \textbf{Google} & 8734 & 8413 & 7055 & \color{condRed}1.5\% & \color{condGreen}0.0\% & \color{condBlue}1.5\% & \color{condRed}1.2\% & \color{condGreen}3.7\% & \color{condBlue}4.9\% & \color{condRed}0.7\% & \color{condGreen}19.2\% & \color{condBlue}19.9\% & Appendix A.5.2 \\
 & & & \textbf{Goodreads} & 0 & 0 & 0 & \color{condRed}-- & \color{condGreen}-- & \color{condBlue}-- & \color{condRed}-- & \color{condGreen}-- & \color{condBlue}0.0\% & \color{condRed}-- & \color{condGreen}-- & \color{condBlue}-- & No Datapoints \\
\midrule

\multirow{3}{*}{\textbf{Scenario 6}} & \multirow{3}{*}{\textbf{MEDIAN}} & \multirow{3}{*}{\textbf{0--50}} 
 & \textbf{Amazon} & 35525 & 34045 & 25483 & \color{condRed}3.2\% & \color{condGreen}0.0\% & \color{condBlue}3.2\% & \color{condRed}3.1\% & \color{condGreen}4.2\% & \color{condBlue}7.3\% & \color{condRed}1.9\% & \color{condGreen}28.3\% & \color{condBlue}30.2\% & Appendix A.6.1 \\
 & & & \textbf{Google} & 13934 & 13369 & 11148 & \color{condRed}1.5\% & \color{condGreen}0.0\% & \color{condBlue}1.5\% & \color{condRed}1.4\% & \color{condGreen}4.1\% & \color{condBlue}5.5\% & \color{condRed}0.5\% & \color{condGreen}20.0\% & \color{condBlue}20.5\% & Appendix A.6.2 \\
 & & & \textbf{Goodreads} & 45474 & 41306 & 32666 & \color{condRed}4.0\% & \color{condGreen}0.0\% & \color{condBlue}4.0\% & \color{condRed}1.3\% & \color{condGreen}9.2\% & \color{condBlue}10.4\% & \color{condRed}1.2\% & \color{condGreen}28.2\% & \color{condBlue}29.4\% & Appendix A.6.3 \\
\bottomrule
\end{tabularx}

\vspace{8pt}
\caption{Consistency Improvements for SSAS vs. DIRECT across diverse dataset scenarios}
\label{tab:consistency_improvements}
\end{table*}

\subsubsection{Base Scenario Performance}
\begin{figure}[htbp] 
    \centering
    \captionsetup{justification=centering} 
    
    \includegraphics[width=\textwidth, height=7.5cm, keepaspectratio=false]{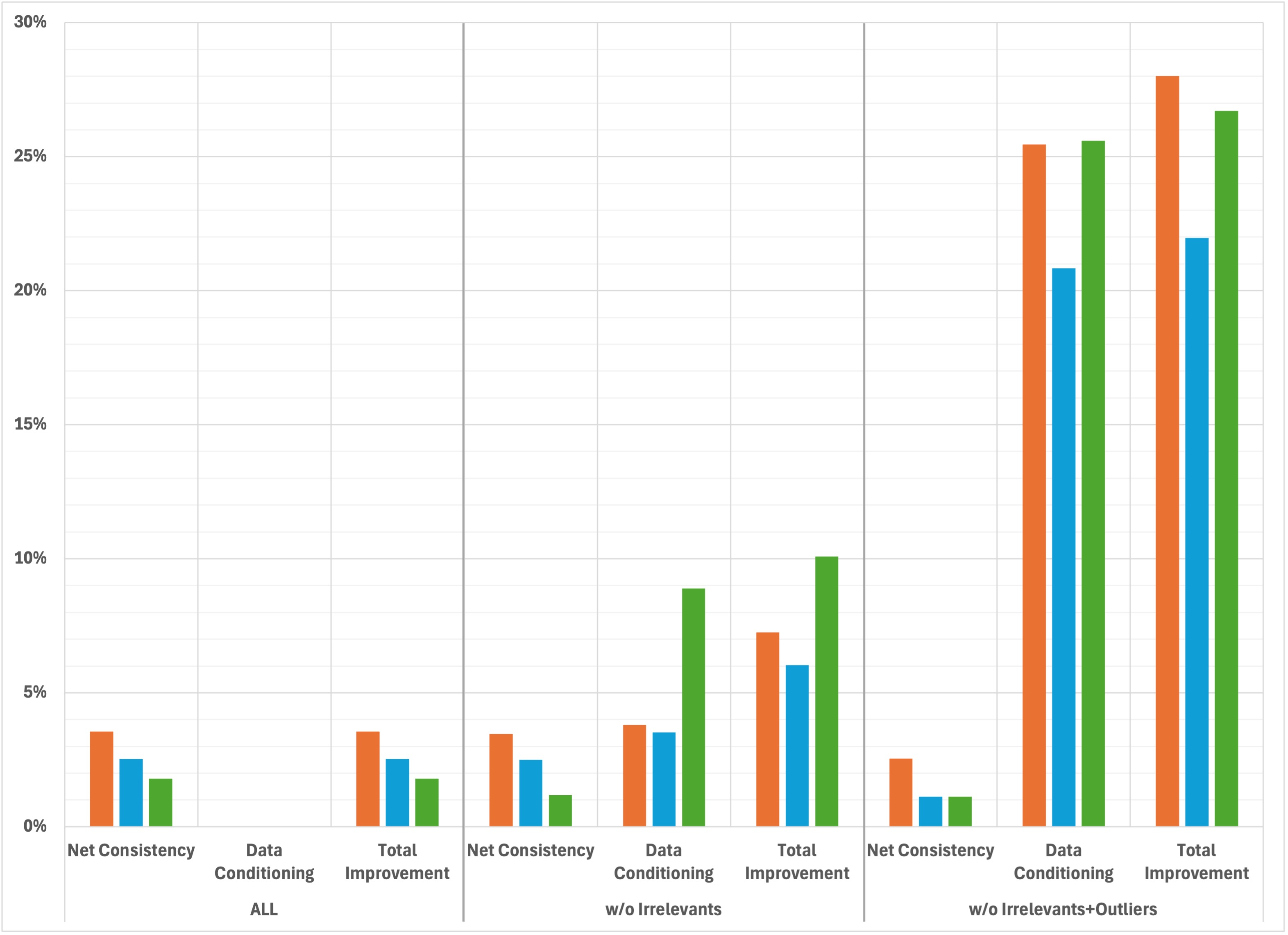} 
    
    \caption{Output Consistency Improvement Overview for BASE Case  (Amazon Product Reviews are in RED, Google Business Reviews are in BLUE, and Goodreads Book Reviews are in GREEN)}
    \label{fig:base consistency improvement}
\end{figure}

As shown in Figure \ref{fig:base consistency improvement}, our methodology demonstrated significant performance gains in the Base Scenario, exceeding the DIRECT baseline consistency by 22–28\% across the three datasets. This improvement is driven by three primary factors:
\begin{enumerate}
    \item Net Consistency improvements ranging from 1.1\% to 2.5\%
    \item Data Conditioning improvements ranging from 20.8\% to 25.6\%, underscoring the high value of our noise-reduction protocol
\end{enumerate}
These improvement patterns are materially identical in all three datasets that we have examined.

\subsubsection{Robustness and Scalability}
Our methodology maintained a high level of performance, across the six robustness scenarios, consistently yielding ~20\% improvement over the baseline. Detailed evidence for specific scenario cases and their corresponding data distributions can be found in Appendix A as noted in the "Supporting Evidence" column of Table \ref{tab:consistency_improvements}. An essential observation derived from our robustness testing is the high degree of data concentration within a minority of primary entities across all three platforms. Our analysis reveals a long-tail distribution where a small percentage of entities account for the vast majority of review volume:
\begin{itemize}
    \item Amazon Product Reviews: 16,395 unique stores were analyzed, totaling 155,745 reviews. However, 77.2\% of the total review volume was generated by only 19\% of the stores.
    \item Google Business Reviews: 251 businesses generated 121,826 reviews, with 81.4\% of those reviews concentrated in just 20.3\% of the businesses.
    \item Goodreads Book Reviews: Out of 19,725 book titles and 157,407 reviews, 71.1\% of the engagement was directed at only 26.4\% of the titles.
\end{itemize}

This concentration highlights the necessity of our Normalized Volume and Review Distribution controls, as they ensure that the methodology remains robust even when faced with significant data skewness and varying levels of entity engagement.

\section{Conclusions}
We addressed a core problem in today’s large-scale text processing domain i.e. the fundamental conflict between the stochastic nature of current LLM outputs and the consistency requirements of enterprise-grade data analytics, particularly around sentiment prediction. This conflict manifests as an Inconsistency Problem, where the same data can yield different sentiment results, and a Noise Problem from chaotic datasets. Our proposed solution,  Syntactic \& Semantic Context Assessment Summarization (SSAS) Methodology, overcomes these issues by implementing a bounded attention mechanism through two phases: assessing Contextual Relevancy and performing Noise Reduction. It uses a structured Hierarchical Data Classification (Themes, Stories, Clusters) and a Summary-of-Summaries (SoS) to pre-process data into a sentiment-dense narrative, forcing the LLM to focus on high-signal tokens. Our evaluation results showed that our SSAS methodology was able to significantly improve data quality by up to 30\% for the three datasets. The SSAS methodology's primary value lies in its ability to enforce a standard of consistency, transforming LLMs from creative tools into reliable analytical instruments. Empirical results from testing SSAS against a direct LLM approach across three diverse datasets confirmed its capacity for aggressive noise mitigation and effective data conditioning through the filtering of irrelevant and outlier data. The business implications are significant - we've implemented SSAS-based prediction mechanisms in multiple domains including marketing \cite{kathuria_leveraging_2026} and supply chain analytics \cite{joglekar_ai_2025}. SSAS mitigates strategic risk by turning volatile sentiment predictions into a stable, repeatable foundation for C-suite decision-making. By alleviating the technical burden of engineering deterministic outputs, it empowers text processors, such as marketing teams, to focus on brand strategy and consumer behavior, ultimately leading to actionable, high-quality insights from large-scale, messy data.

\clearpage 
\bibliographystyle{unsrturl}
\bibliography{references}

\clearpage
\appendix

\renewcommand{\thetable}{\thesection.\arabic{subsection}.\arabic{table}}
\setcounter{table}{0}
\counterwithin*{table}{subsection} 


\section{Base Case}
\subsection{Amazon Product Reviews}

\begin{table*}[h!]
    \centering
    \scriptsize
    \renewcommand{\arraystretch}{1.3}
    \setlength{\tabcolsep}{1.1pt} 

\begin{tabularx}{\textwidth}{l *{5}{>{\centering\arraybackslash}X} @{\hspace{3pt}} *{5}{>{\centering\arraybackslash}X} @{\hspace{3pt}} *{5}{>{\centering\arraybackslash}X}}
    \toprule
     & \multicolumn{5}{c}{\textbf{ALL Data}} & \multicolumn{5}{c}{\textbf{w/o Irrelevant}} & \multicolumn{5}{c}{\textbf{w/o Irrelevant, Outlier}} \\
    \cmidrule(lr){2-6} \cmidrule(lr){7-11} \cmidrule(lr){12-16}
     & \multicolumn{3}{c}{\textbf{100\% Consistent}} & \textbf{Inc.} & \textbf{Total} & \multicolumn{3}{c}{\textbf{100\% Consistent}} & \textbf{Inc.} & \textbf{Total} & \multicolumn{3}{c}{\textbf{100\% Consistent}} & \textbf{Inc.} & \textbf{Total} \\
    \cmidrule(lr){2-4} \cmidrule(lr){7-9} \cmidrule(lr){12-14}
    \textbf{Method} & \textbf{Pos.} & \textbf{Neg.} & \textbf{Neu.} & \textbf{0--99\%} & & \textbf{Pos.} & \textbf{Neg.} & \textbf{Neu.} & \textbf{0--99\%} & & \textbf{Pos.} & \textbf{Neg.} & \textbf{Neu.} & \textbf{0--99\%} & \\ 
    \midrule

    DIRECT & 84,938 & 43,279 & 7,634 & 19,894 & 155,745 & 81,810 & 41,654 & 6,929 & 19,430 & 149,823 & 65,416 & 30,903 & 3,762 & 16,021 & 116,102 \\
    SSAS & 86,861 & 44,122 & 7,125 & 17,637 & 155,745 & 83,565 & 42,489 & 6,475 & 17,294 & 149,823 & 66,340 & 31,459 & 3,580 & 14,723 & 116,102 \\
    \textbf{Diff} & 1,923 & 843 & --509 & --2,257 & 5,532 & 1,755 & 835 & --454 & --2,136 & 5,180 & 924 & 556 & --182 & --1,298 & 2,960 \\ 
    \midrule
    DIRECT \% & 54.54\% & 27.79\% & 4.90\% & 12.77\% & 100\% & 54.60\% & 27.80\% & 4.62\% & 12.97\% & 100\% & 56.34\% & 26.62\% & 3.24\% & 13.80\% & 100\% \\
    SSAS \% & 55.77\% & 28.33\% & 4.57\% & 11.32\% & 100\% & 55.78\% & 28.36\% & 4.32\% & 11.54\% & 100\% & 57.14\% & 27.10\% & 3.08\% & 12.68\% & 100\% \\
\bottomrule
    \end{tabularx}
    \vspace{8pt}
    \caption{Detailed Consistency Metrics for Amazon Product Reviews (Base Case).}
    \label{tab:appx_amazon_base}
\end{table*}

\begin{table}[H]
\centering
\scriptsize
\renewcommand{\arraystretch}{1.3}
\setlength{\tabcolsep}{1pt}

\begin{tabularx}{\textwidth}{l l *{5}{>{\centering\arraybackslash}X} @{\hspace{2pt}} *{5}{>{\centering\arraybackslash}X} @{\hspace{2pt}} *{5}{>{\centering\arraybackslash}X}}
\toprule
 & & \multicolumn{15}{c}{\textbf{SSAS}} \\
 \cmidrule(lr){3-17}
 & & \multicolumn{5}{c}{\textbf{ALL Data}} & \multicolumn{5}{c}{\textbf{w/o Irrelevant}} & \multicolumn{5}{c}{\textbf{w/o Irrelevant, Outlier}} \\
\cmidrule(lr){3-7} \cmidrule(lr){8-12} \cmidrule(lr){13-17}
\textbf{Source} & \textbf{Datapoints} & \textbf{Pos.} & \textbf{Neg.} & \textbf{Neu.} & \textbf{Inc.} & \textbf{Total} & \textbf{Pos.} & \textbf{Neg.} & \textbf{Neu.} & \textbf{Inc.} & \textbf{Total} & \textbf{Pos.} & \textbf{Neg.} & \textbf{Neu.} & \textbf{Inc.} & \textbf{Total} \\
\midrule

\multirow{4}{*}{\rotatebox[origin=c]{90}{\textbf{DIRECT}}} 
& \textbf{Positive} & 82,192 & \textcolor{orange}{88} & \textcolor{orange}{194} & \textcolor{red}{2,464} & 84,938 & 79,173 & \textcolor{orange}{78} & \textcolor{orange}{161} & \textcolor{red}{2,398} & 81,810 & 63,465 & \textcolor{orange}{2} & \textcolor{orange}{8} & \textcolor{red}{1,941} & 65,416 \\

& \textbf{Negative} & \textcolor{orange}{117} & 41,705 & \textcolor{orange}{298} & \textcolor{red}{1,159} & 43,279 & \textcolor{orange}{101} & 40,177 & \textcolor{orange}{257} & \textcolor{red}{1,119} & 41,654 & \textcolor{orange}{0} & 30,084 & \textcolor{orange}{4} & \textcolor{red}{815} & 30,903 \\

& \textbf{Neutral} & \textcolor{orange}{605} & \textcolor{orange}{237} & 5,211 & \textcolor{red}{1,581} & 7,634 & \textcolor{orange}{525} & \textcolor{orange}{209} & 4,687 & \textcolor{red}{1,508} & 6,929 & \textcolor{orange}{119} & \textcolor{orange}{4} & 2,635 & \textcolor{red}{1,004} & 3,762 \\
\bottomrule
\end{tabularx}

\vspace{8pt}
\caption{Confusion Matrix (Amazon Product Reviews): Base Case across refinement stages.}
\label{tab:appx_cm_amazon_base}
\end{table}

\subsection{Google Business Reviews}

\begin{table*}[h!]
    \centering
    \scriptsize
    \renewcommand{\arraystretch}{1.3}
    \setlength{\tabcolsep}{1.1pt}
    \begin{tabularx}{\textwidth}{l *{5}{>{\centering\arraybackslash}X} @{\hspace{3pt}} *{5}{>{\centering\arraybackslash}X} @{\hspace{3pt}} *{5}{>{\centering\arraybackslash}X}}
    \toprule
     & \multicolumn{5}{c}{\textbf{ALL Data}} & \multicolumn{5}{c}{\textbf{w/o Irrelevant}} & \multicolumn{5}{c}{\textbf{w/o Irrelevant, Outlier}} \\
    \cmidrule(lr){2-6} \cmidrule(lr){7-11} \cmidrule(lr){12-16}
     & \multicolumn{3}{c}{\textbf{100\% Consistent}} & \textbf{Inc.} & \textbf{Total} & \multicolumn{3}{c}{\textbf{100\% Consistent}} & \textbf{Inc.} & \textbf{Total} & \multicolumn{3}{c}{\textbf{100\% Consistent}} & \textbf{Inc.} & \textbf{Total} \\
    \cmidrule(lr){2-4} \cmidrule(lr){7-9} \cmidrule(lr){12-14}
    \textbf{Method} & \textbf{Pos.} & \textbf{Neg.} & \textbf{Neu.} & \textbf{0--99\%} & & \textbf{Pos.} & \textbf{Neg.} & \textbf{Neu.} & \textbf{0--99\%} & & \textbf{Pos.} & \textbf{Neg.} & \textbf{Neu.} & \textbf{0--99\%} & \\ 
    \midrule
    DIRECT & 74,859 & 27,664 & 6,155 & 13,148 & 121,826 & 72,725 & 26,441 & 5,636 & 12,726 & 117,528 & 63,213 & 20,036 & 3,143 & 10,042 & 96,434 \\
    SSAS & 75,731 & 28,330 & 5,779 & 11,986 & 121,826 & 73,568 & 27,070 & 5,302 & 11,588 & 117,528 & 63,483 & 20,308 & 3,079 & 9,564 & 96,434 \\
    \textbf{Diff} & 872 & 666 & --376 & --1,162 & 3,076 & 843 & 629 & --334 & --1,138 & 2,944 & 270 & 272 & --64 & --478 & 1,084 \\ 
    \midrule
    DIRECT \% & 61.45\% & 22.71\% & 5.05\% & 10.79\% & 100\% & 61.88\% & 22.50\% & 4.80\% & 10.83\% & 100\% & 65.55\% & 20.78\% & 3.26\% & 10.41\% & 100\% \\
    SSAS \% & 62.16\% & 23.25\% & 4.74\% & 9.84\% & 100\% & 62.60\% & 23.03\% & 4.51\% & 9.86\% & 100\% & 65.83\% & 21.06\% & 3.19\% & 9.92\% & 100\% \\
\bottomrule
    \end{tabularx}
    \vspace{8pt}
    \caption{Detailed Consistency Metrics for Google Business Reviews (Base Case).}
\end{table*}

\begin{table}[H]
\centering
\scriptsize
\renewcommand{\arraystretch}{1.3}
\setlength{\tabcolsep}{1pt}

\begin{tabularx}{\textwidth}{l l *{5}{>{\centering\arraybackslash}X} @{\hspace{2pt}} *{5}{>{\centering\arraybackslash}X} @{\hspace{2pt}} *{5}{>{\centering\arraybackslash}X}}
\toprule
 & & \multicolumn{15}{c}{\textbf{SSAS}} \\
 \cmidrule(lr){3-17}
 & & \multicolumn{5}{c}{\textbf{ALL Data}} & \multicolumn{5}{c}{\textbf{w/o Irrelevant}} & \multicolumn{5}{c}{\textbf{w/o Irrelevant, Outlier}} \\
\cmidrule(lr){3-7} \cmidrule(lr){8-12} \cmidrule(lr){13-17}
\textbf{Source} & \textbf{Datapoints} & \textbf{Pos.} & \textbf{Neg.} & \textbf{Neu.} & \textbf{Inc.} & \textbf{Total} & \textbf{Pos.} & \textbf{Neg.} & \textbf{Neu.} & \textbf{Inc.} & \textbf{Total} & \textbf{Pos.} & \textbf{Neg.} & \textbf{Neu.} & \textbf{Inc.} & \textbf{Total} \\
\midrule

\multirow{4}{*}{\rotatebox[origin=c]{90}{\textbf{DIRECT}}} 
& \textbf{Positive} & 73,270 & \textcolor{orange}{35} & \textcolor{orange}{130} & \textcolor{red}{1,424} & 74,859 & 71,241 & \textcolor{orange}{31} & \textcolor{orange}{104} & \textcolor{red}{1,349} & 72,725 & 62,115 & \textcolor{orange}{1} & \textcolor{orange}{2} & \textcolor{red}{1,095} & 63,213 \\

& \textbf{Negative} & \textcolor{orange}{48} & 26,726 & \textcolor{orange}{121} & \textcolor{red}{769} & 27,664 & \textcolor{orange}{43} & 25,564 & \textcolor{orange}{99} & \textcolor{red}{735} & 26,441 & \textcolor{orange}{3} & 19,484 & \textcolor{orange}{0} & \textcolor{red}{549} & 20,036 \\

& \textbf{Neutral} & \textcolor{orange}{296} & \textcolor{orange}{143} & 4,708 & \textcolor{red}{1,008} & 6,155 & \textcolor{orange}{248} & \textcolor{orange}{121} & 4,331 & \textcolor{red}{936} & 5,636 & \textcolor{orange}{15} & \textcolor{orange}{0} & 2,608 & \textcolor{red}{520} & 3,143 \\
\bottomrule
\end{tabularx}

\vspace{8pt}
\caption{Confusion Matrix (Google Business Reviews): Base Case across refinement stages.}
\label{tab:appx_cm_google_base}
\end{table}

\subsection{Goodreads Book Reviews}

\begin{table*}[h!]
    \centering
    \scriptsize
    \renewcommand{\arraystretch}{1.3}
    \setlength{\tabcolsep}{1.1pt}
    \begin{tabularx}{\textwidth}{l *{5}{>{\centering\arraybackslash}X} @{\hspace{3pt}} *{5}{>{\centering\arraybackslash}X} @{\hspace{3pt}} *{5}{>{\centering\arraybackslash}X}}
    \toprule
     & \multicolumn{5}{c}{\textbf{ALL Data}} & \multicolumn{5}{c}{\textbf{w/o Irrelevant}} & \multicolumn{5}{c}{\textbf{w/o Irrelevant, Outlier}} \\
    \cmidrule(lr){2-6} \cmidrule(lr){7-11} \cmidrule(lr){12-16}
     & \multicolumn{3}{c}{\textbf{100\% Consistent}} & \textbf{Inc.} & \textbf{Total} & \multicolumn{3}{c}{\textbf{100\% Consistent}} & \textbf{Inc.} & \textbf{Total} & \multicolumn{3}{c}{\textbf{100\% Consistent}} & \textbf{Inc.} & \textbf{Total} \\
    \cmidrule(lr){2-4} \cmidrule(lr){7-9} \cmidrule(lr){12-14}
    \textbf{Method} & \textbf{Pos.} & \textbf{Neg.} & \textbf{Neu.} & \textbf{0--99\%} & & \textbf{Pos.} & \textbf{Neg.} & \textbf{Neu.} & \textbf{0--99\%} & & \textbf{Pos.} & \textbf{Neg.} & \textbf{Neu.} & \textbf{0--99\%} & \\ 
    \midrule
    DIRECT & 81918 & 31001 & 8491 & 35997 & 157407 & 75000 & 27694 & 6998 & 33708 & 143400 & 59509 & 22806 & 4308 & 30510 & 117133 \\
    SSAS & 82721 & 31605 & 7410 & 35671 & 157407 & 75453 & 28088 & 6162 & 33697 & 143400 & 58962 & 23079 & 4194 & 30898 & 117133 \\
    \textbf{Diff} & 803 & 604 & 1081 & 326 & 2814 & 453 & 394 & 836 & 11 & 1694 & 547 & 273 & 114 & 388 & 1322 \\ 
    \midrule
    DIRECT \% & 52.04\% & 19.69\% & 5.39\% & 22.87\% & 100\% & 52.30\% & 19.31\% & 4.88\% & 23.51\% & 100\% & 50.80\% & 19.47\% & 3.68\% & 26.05\% & 100\% \\
    SSAS \% & 52.55\% & 20.08\% & 4.71\% & 22.66\% & 100\% & 52.62\% & 19.59\% & 4.30\% & 23.50\% & 100\% & 50.34\% & 19.70\% & 3.58\% & 26.38\% & 100\% \\
\bottomrule
    \end{tabularx}
    \vspace{8pt}
    \caption{Detailed Consistency Metrics for Goodreads Book Reviews (Base Case).}
\end{table*}

\begin{table}[H]
\centering
\scriptsize
\renewcommand{\arraystretch}{1.3}
\setlength{\tabcolsep}{1pt}

\begin{tabularx}{\textwidth}{l l *{5}{>{\centering\arraybackslash}X} @{\hspace{2pt}} *{5}{>{\centering\arraybackslash}X} @{\hspace{2pt}} *{5}{>{\centering\arraybackslash}X}}
\toprule
 & & \multicolumn{15}{c}{\textbf{SSAS}} \\
 \cmidrule(lr){3-17}
 & & \multicolumn{5}{c}{\textbf{ALL Data}} & \multicolumn{5}{c}{\textbf{w/o Irrelevant}} & \multicolumn{5}{c}{\textbf{w/o Irrelevant, Outlier}} \\
\cmidrule(lr){3-7} \cmidrule(lr){8-12} \cmidrule(lr){13-17}
\textbf{Source} & \textbf{Datapoints} & \textbf{Pos.} & \textbf{Neg.} & \textbf{Neu.} & \textbf{Inc.} & \textbf{Total} & \textbf{Pos.} & \textbf{Neg.} & \textbf{Neu.} & \textbf{Inc.} & \textbf{Total} & \textbf{Pos.} & \textbf{Neg.} & \textbf{Neu.} & \textbf{Inc.} & \textbf{Total} \\
\midrule

\multirow{4}{*}{\rotatebox[origin=c]{90}{\textbf{DIRECT}}} 
& \textbf{Positive} & 77,510 & \textcolor{orange}{139} & \textcolor{orange}{115} & \textcolor{red}{4,154} & 81,918 & 70,950 & \textcolor{orange}{101} & \textcolor{orange}{66} & \textcolor{red}{3,883} & 75,000 & 56,046 & \textcolor{orange}{0} & \textcolor{orange}{1} & \textcolor{red}{3,462} & 59,509 \\

& \textbf{Negative} & \textcolor{orange}{312} & 28,230 & \textcolor{orange}{138} & \textcolor{red}{2,321} & 31,001 & \textcolor{orange}{209} & 25,216 & \textcolor{orange}{94} & \textcolor{red}{2,175} & 27,694 & \textcolor{orange}{0} & 20,974 & \textcolor{orange}{0} & \textcolor{red}{1,832} & 22,806 \\

& \textbf{Neutral} & \textcolor{orange}{597} & \textcolor{orange}{259} & 5,341 & \textcolor{red}{2,294} & 8,491 & \textcolor{orange}{465} & \textcolor{orange}{189} & 4,367 & \textcolor{red}{1,977} & 6,998 & \textcolor{orange}{33} & \textcolor{orange}{2} & 2,822 & \textcolor{red}{1,451} & 4,308 \\
\bottomrule
\end{tabularx}

\vspace{8pt}
\caption{Confusion Matrix (Goodreads Book Reviews): Base Case across refinement stages.}
\label{tab:appx_cm_goodreads_base}
\end{table}

\clearpage
\section{Scenario 1 – HIGH Volume, 100\% Review Distribution}
\subsection{Amazon Product Reviews}

\begin{table*}[h!]
    \centering
    \scriptsize
    \renewcommand{\arraystretch}{1.3}
    \setlength{\tabcolsep}{1.1pt}
    \begin{tabularx}{\textwidth}{l *{5}{>{\centering\arraybackslash}X} @{\hspace{3pt}} *{5}{>{\centering\arraybackslash}X} @{\hspace{3pt}} *{5}{>{\centering\arraybackslash}X}}
    \toprule
     & \multicolumn{5}{c}{\textbf{ALL Data}} & \multicolumn{5}{c}{\textbf{w/o Irrelevant}} & \multicolumn{5}{c}{\textbf{w/o Irrelevant, Outlier}} \\
    \cmidrule(lr){2-6} \cmidrule(lr){7-11} \cmidrule(lr){12-16}
     & \multicolumn{3}{c}{\textbf{100\% Consistent}} & \textbf{Inc.} & \textbf{Total} & \multicolumn{3}{c}{\textbf{100\% Consistent}} & \textbf{Inc.} & \textbf{Total} & \multicolumn{3}{c}{\textbf{100\% Consistent}} & \textbf{Inc.} & \textbf{Total} \\
    \cmidrule(lr){2-4} \cmidrule(lr){7-9} \cmidrule(lr){12-14}
    \textbf{Method} & \textbf{Pos.} & \textbf{Neg.} & \textbf{Neu.} & \textbf{0--99\%} & & \textbf{Pos.} & \textbf{Neg.} & \textbf{Neu.} & \textbf{0--99\%} & & \textbf{Pos.} & \textbf{Neg.} & \textbf{Neu.} & \textbf{0--99\%} & \\ 
    \midrule
    DIRECT & 9,353 & 4,780 & 867 & 2,131 & 17,131 & 8,985 & 4,615 & 775 & 2,067 & 16,442 & 7,225 & 3,369 & 423 & 1,629 & 12,646 \\
    SSAS & 9,605 & 4,882 & 789 & 1,855 & 17,131 & 9,210 & 4,714 & 708 & 1,810 & 16,442 & 7,351 & 3,433 & 373 & 1,489 & 12,646 \\
    \textbf{Diff} & 252 & 102 & 78 & 276 & 708 & 225 & 99 & 67 & 257 & 648 & 126 & 64 & 50 & 140 & 380 \\ 
    \midrule
    DIRECT \% & 54.60\% & 27.90\% & 5.06\% & 12.44\% & 100\% & 54.65\% & 28.07\% & 4.71\% & 12.57\% & 100\% & 57.13\% & 26.64\% & 3.34\% & 12.88\% & 100\% \\
    SSAS \% & 56.07\% & 28.50\% & 4.61\% & 10.83\% & 100\% & 56.02\% & 28.67\% & 4.31\% & 11.01\% & 100\% & 58.13\% & 27.15\% & 2.95\% & 11.77\% & 100\% \\
\bottomrule
    \end{tabularx}
    \vspace{8pt}
    \caption{Detailed Consistency Metrics for Amazon Product Reviews (Scenario 1).}
    \label{tab:appx_amazon_s1}
\end{table*}

\begin{table}[H]
\centering
\scriptsize
\renewcommand{\arraystretch}{1.3}
\setlength{\tabcolsep}{1pt}

\begin{tabularx}{\textwidth}{l l *{5}{>{\centering\arraybackslash}X} @{\hspace{2pt}} *{5}{>{\centering\arraybackslash}X} @{\hspace{2pt}} *{5}{>{\centering\arraybackslash}X}}
\toprule
 & & \multicolumn{15}{c}{\textbf{SSAS}} \\
 \cmidrule(lr){3-17}
 & & \multicolumn{5}{c}{\textbf{ALL Data}} & \multicolumn{5}{c}{\textbf{w/o Irrelevant}} & \multicolumn{5}{c}{\textbf{w/o Irrelevant, Outlier}} \\
\cmidrule(lr){3-7} \cmidrule(lr){8-12} \cmidrule(lr){13-17}
\textbf{Source} & \textbf{Datapoints} & \textbf{Pos.} & \textbf{Neg.} & \textbf{Neu.} & \textbf{Inc.} & \textbf{Total} & \textbf{Pos.} & \textbf{Neg.} & \textbf{Neu.} & \textbf{Inc.} & \textbf{Total} & \textbf{Pos.} & \textbf{Neg.} & \textbf{Neu.} & \textbf{Inc.} & \textbf{Total} \\
\midrule

\multirow{4}{*}{\rotatebox[origin=c]{90}{\textbf{DIRECT}}} 
& \textbf{Positive} & 9,047 & \textcolor{orange}{5} & \textcolor{orange}{23} & \textcolor{red}{278} & 9,353 & 8,691 & \textcolor{orange}{4} & \textcolor{orange}{21} & \textcolor{red}{269} & 8,985 & 7,015 & \textcolor{orange}{0} & \textcolor{orange}{1} & \textcolor{red}{209} & 7,225 \\

& \textbf{Negative} & \textcolor{orange}{12} & 4,614 & \textcolor{orange}{35} & \textcolor{red}{119} & 4,780 & \textcolor{orange}{11} & 4,458 & \textcolor{orange}{30} & \textcolor{red}{116} & 4,615 & \textcolor{orange}{0} & 3,285 & \textcolor{orange}{1} & \textcolor{red}{83} & 3,369 \\

& \textbf{Neutral} & \textcolor{orange}{65} & \textcolor{orange}{22} & 568 & \textcolor{red}{212} & 867 & \textcolor{orange}{51} & \textcolor{orange}{19} & 502 & \textcolor{red}{203} & 775 & \textcolor{orange}{10} & \textcolor{orange}{2} & 273 & \textcolor{red}{138} & 423 \\
\bottomrule
\end{tabularx}

\vspace{8pt}
\caption{Confusion Matrix (Amazon Product Reviews): Scenario 1 (HIGH Volume, 100\% Dist.)}
\label{tab:appx_cm_amazon_s1}
\end{table}

\subsection{Google Business Reviews}
No Data for this combination of Dataset and Scenario

\subsection{Goodreads Book Reviews}
No Data for this combination of Dataset and Scenario

\clearpage
\section{Scenario 2 – HIGH Volume, 51--99\% Review Distribution}
\subsection{Amazon Product Reviews}

\begin{table*}[h!]
    \centering
    \scriptsize
    \renewcommand{\arraystretch}{1.3}
    \setlength{\tabcolsep}{1.1pt}
\begin{tabularx}{\textwidth}{l *{5}{>{\centering\arraybackslash}X} @{\hspace{3pt}} *{5}{>{\centering\arraybackslash}X} @{\hspace{3pt}} *{5}{>{\centering\arraybackslash}X}}
    \toprule
     & \multicolumn{5}{c}{\textbf{ALL Data}} & \multicolumn{5}{c}{\textbf{w/o Irrelevant}} & \multicolumn{5}{c}{\textbf{w/o Irrelevant, Outlier}} \\
    \cmidrule(lr){2-6} \cmidrule(lr){7-11} \cmidrule(lr){12-16}
    \textbf{Method} & \textbf{Pos.} & \textbf{Neg.} & \textbf{Neu.} & \textbf{Inc.} & \textbf{Total} & \textbf{Pos.} & \textbf{Neg.} & \textbf{Neu.} & \textbf{Inc.} & \textbf{Total} & \textbf{Pos.} & \textbf{Neg.} & \textbf{Neu.} & \textbf{Inc.} & \textbf{Total} \\ 
    \midrule
    DIRECT & 33,977 & 15,452 & 3,010 & 7,419 & 59,858 & 32,655 & 14,827 & 2,700 & 7,222 & 57,404 & 26,167 & 10,955 & 1,403 & 5,916 & 44,441 \\
    SSAS & 34,810 & 15,759 & 2,766 & 6,523 & 59,858 & 33,423 & 15,137 & 2,481 & 6,363 & 57,404 & 26,625 & 11,182 & 1,302 & 5,332 & 44,441 \\
    \textbf{Diff} & 833 & 307 & 244 & 896 & 2,280 & 768 & 310 & 219 & 859 & 2,156 & 458 & 227 & 101 & 584 & 1,370 \\ 
    \midrule
    DIRECT \% & 56.76\% & 25.81\% & 5.03\% & 12.39\% & 100\% & 56.89\% & 25.83\% & 4.70\% & 12.58\% & 100\% & 58.88\% & 24.65\% & 3.16\% & 13.31\% & 100\% \\
    SSAS \% & 58.15\% & 26.33\% & 4.62\% & 10.90\% & 100\% & 58.22\% & 26.37\% & 4.32\% & 11.08\% & 100\% & 59.91\% & 25.16\% & 2.93\% & 12.00\% & 100\% \\
    \% Diff & 1.39\% & 0.51\% & 0.41\% & 1.50\% & 3.81\% & 1.34\% & 0.54\% & 0.38\% & 1.50\% & 3.76\% & 1.03\% & 0.51\% & 0.23\% & 1.31\% & 3.08\% \\ 
    \midrule
    \textbf{Improvement} & \multicolumn{2}{c}{\textcolor{blue}{3.81\%}} & \multicolumn{2}{c}{\textcolor{blue}{0.00\%}} & \textcolor{blue}{3.81\%} & \multicolumn{2}{c}{\textcolor{blue}{3.76\%}} & \multicolumn{2}{c}{\textcolor{blue}{4.10\%}} & \textcolor{blue}{7.86\%} & \multicolumn{2}{c}{\textcolor{blue}{3.08\%}} & \multicolumn{2}{c}{\textcolor{blue}{25.76\%}} & \textcolor{blue}{28.84\%} \\
    \bottomrule
    \end{tabularx}
    \vspace{8pt}
    \caption{Detailed Consistency Metrics for Amazon Product Reviews (Scenario 2).}
    \label{tab:appx_amazon_s2}
\end{table*}

\begin{table}[H]
\centering
\scriptsize
\renewcommand{\arraystretch}{1.3}
\setlength{\tabcolsep}{1pt}

\begin{tabularx}{\textwidth}{l l *{5}{>{\centering\arraybackslash}X} @{\hspace{2pt}} *{5}{>{\centering\arraybackslash}X} @{\hspace{2pt}} *{5}{>{\centering\arraybackslash}X}}
\toprule
 & & \multicolumn{15}{c}{\textbf{SSAS}} \\
 \cmidrule(lr){3-17}
 & & \multicolumn{5}{c}{\textbf{ALL Data}} & \multicolumn{5}{c}{\textbf{w/o Irrelevant}} & \multicolumn{5}{c}{\textbf{w/o Irrelevant, Outlier}} \\
\cmidrule(lr){3-7} \cmidrule(lr){8-12} \cmidrule(lr){13-17}
\textbf{Source} & \textbf{Datapoints} & \textbf{Pos.} & \textbf{Neg.} & \textbf{Neu.} & \textbf{Inc.} & \textbf{Total} & \textbf{Pos.} & \textbf{Neg.} & \textbf{Neu.} & \textbf{Inc.} & \textbf{Total} & \textbf{Pos.} & \textbf{Neg.} & \textbf{Neu.} & \textbf{Inc.} & \textbf{Total} \\
\midrule

\multirow{4}{*}{\rotatebox[origin=c]{90}{\textbf{DIRECT}}} 
& \textbf{Positive} & 32,900 & \textcolor{orange}{43} & \textcolor{orange}{82} & \textcolor{red}{952} & 33,977 & 31,626 & \textcolor{orange}{38} & \textcolor{orange}{69} & \textcolor{red}{922} & 32,655 & 25,435 & \textcolor{orange}{2} & \textcolor{orange}{1} & \textcolor{red}{729} & 26,167 \\

& \textbf{Negative} & \textcolor{orange}{47} & 14,876 & \textcolor{orange}{110} & \textcolor{red}{419} & 15,452 & \textcolor{orange}{38} & 14,296 & \textcolor{orange}{94} & \textcolor{red}{399} & 14,827 & \textcolor{orange}{0} & 10,676 & \textcolor{orange}{1} & \textcolor{red}{278} & 10,955 \\

& \textbf{Neutral} & \textcolor{orange}{260} & \textcolor{orange}{91} & 2,054 & \textcolor{red}{605} & 3,010 & \textcolor{orange}{229} & \textcolor{orange}{77} & 1,826 & \textcolor{red}{568} & 2,700 & \textcolor{orange}{63} & \textcolor{orange}{2} & 973 & \textcolor{red}{365} & 1,403 \\
\bottomrule
\end{tabularx}

\vspace{8pt}
\caption{Confusion Matrix (Amazon Product Reviews): Scenario 2 (HIGH Volume, 51--99\% Dist.)}
\label{tab:appx_cm_amazon_s2}
\end{table}

\subsection{Google Business Reviews}

\begin{table*}[h!]
    \centering
    \scriptsize
    \renewcommand{\arraystretch}{1.3}
    \setlength{\tabcolsep}{1.1pt}
    \begin{tabularx}{\textwidth}{l *{5}{>{\centering\arraybackslash}X} @{\hspace{3pt}} *{5}{>{\centering\arraybackslash}X} @{\hspace{3pt}} *{5}{>{\centering\arraybackslash}X}}
    \toprule
     & \multicolumn{5}{c}{\textbf{ALL Data}} & \multicolumn{5}{c}{\textbf{w/o Irrelevant}} & \multicolumn{5}{c}{\textbf{w/o Irrelevant, Outlier}} \\
    \cmidrule(lr){2-6} \cmidrule(lr){7-11} \cmidrule(lr){12-16}
    \textbf{Method} & \textbf{Pos.} & \textbf{Neg.} & \textbf{Neu.} & \textbf{Inc.} & \textbf{Total} & \textbf{Pos.} & \textbf{Neg.} & \textbf{Neu.} & \textbf{Inc.} & \textbf{Total} & \textbf{Pos.} & \textbf{Neg.} & \textbf{Neu.} & \textbf{Inc.} & \textbf{Total} \\ 
    \midrule
    DIRECT & 54,788 & 23,143 & 5,199 & 10,510 & 93,640 & 53,290 & 22,129 & 4,790 & 10,169 & 90,378 & 46,100 & 16,789 & 2,696 & 8,027 & 73,612 \\
    SSAS & 55,563 & 23,676 & 4,846 & 9,555 & 93,640 & 54,055 & 22,630 & 4,477 & 9,216 & 90,378 & 46,411 & 16,995 & 2,621 & 7,585 & 73,612 \\
    \textbf{Diff} & 775 & 533 & --353 & --955 & 2,616 & 765 & 501 & --313 & --953 & 2,532 & 311 & 206 & --75 & --442 & 1,034 \\ 
    \midrule
    DIRECT \% & 58.51\% & 24.71\% & 5.55\% & 11.22\% & 100\% & 58.96\% & 24.48\% & 5.30\% & 11.25\% & 100\% & 62.63\% & 22.81\% & 3.66\% & 10.90\% & 100\% \\
    SSAS \% & 59.34\% & 25.28\% & 5.18\% & 10.20\% & 100\% & 59.81\% & 25.04\% & 4.95\% & 10.20\% & 100\% & 63.05\% & 23.09\% & 3.56\% & 10.30\% & 100\% \\
    \% Diff & 0.83\% & 0.57\% & --0.38\% & --1.02\% & 2.79\% & 0.85\% & 0.55\% & --0.35\% & --1.05\% & 2.80\% & 0.42\% & 0.28\% & --0.10\% & --0.60\% & 1.40\% \\
\bottomrule
    \end{tabularx}
    \vspace{8pt}
    \caption{Detailed Consistency Metrics for Google Business Reviews (Scenario 2).}
\end{table*}

\begin{table}[H]
\centering
\scriptsize
\renewcommand{\arraystretch}{1.3}
\setlength{\tabcolsep}{1pt}

\begin{tabularx}{\textwidth}{l l *{5}{>{\centering\arraybackslash}X} @{\hspace{2pt}} *{5}{>{\centering\arraybackslash}X} @{\hspace{2pt}} *{5}{>{\centering\arraybackslash}X}}
\toprule
 & & \multicolumn{15}{c}{\textbf{SSAS}} \\
 \cmidrule(lr){3-17}
 & & \multicolumn{5}{c}{\textbf{ALL Data}} & \multicolumn{5}{c}{\textbf{w/o Irrelevant}} & \multicolumn{5}{c}{\textbf{w/o Irrelevant, Outlier}} \\
\cmidrule(lr){3-7} \cmidrule(lr){8-12} \cmidrule(lr){13-17}
\textbf{Source} & \textbf{Datapoints} & \textbf{Pos.} & \textbf{Neg.} & \textbf{Neu.} & \textbf{Inc.} & \textbf{Total} & \textbf{Pos.} & \textbf{Neg.} & \textbf{Neu.} & \textbf{Inc.} & \textbf{Total} & \textbf{Pos.} & \textbf{Neg.} & \textbf{Neu.} & \textbf{Inc.} & \textbf{Total} \\
\midrule

\multirow{4}{*}{\rotatebox[origin=c]{90}{\textbf{DIRECT}}} 
& \textbf{Positive} & 53,583 & \textcolor{orange}{27} & \textcolor{orange}{95} & \textcolor{red}{1,083} & 54,788 & 52,178 & \textcolor{orange}{24} & \textcolor{orange}{73} & \textcolor{red}{1,015} & 53,290 & 45,272 & \textcolor{orange}{1} & \textcolor{orange}{1} & \textcolor{red}{826} & 46,100 \\

& \textbf{Negative} & \textcolor{orange}{37} & 22,390 & \textcolor{orange}{94} & \textcolor{red}{622} & 23,143 & \textcolor{orange}{32} & 21,420 & \textcolor{orange}{80} & \textcolor{red}{597} & 22,129 & \textcolor{orange}{3} & 16,339 & \textcolor{orange}{0} & \textcolor{red}{447} & 16,789 \\

& \textbf{Neutral} & \textcolor{orange}{224} & \textcolor{orange}{117} & 4,020 & \textcolor{red}{838} & 5,199 & \textcolor{orange}{186} & \textcolor{orange}{99} & 3,729 & \textcolor{red}{776} & 4,790 & \textcolor{orange}{9} & \textcolor{orange}{0} & 2,260 & \textcolor{red}{427} & 2,696 \\
\bottomrule
\end{tabularx}

\vspace{8pt}
\caption{Confusion Matrix (Google Business Reviews): Scenario 2 (HIGH Volume, 51--99\% Dist.)}
\label{tab:appx_cm_google_s2}
\end{table}

\subsection{Goodreads Book Reviews}

\begin{table*}[h!]
    \centering
    \scriptsize
    \renewcommand{\arraystretch}{1.3}
    \setlength{\tabcolsep}{1.1pt}
    \begin{tabularx}{\textwidth}{l *{5}{>{\centering\arraybackslash}X} @{\hspace{3pt}} *{5}{>{\centering\arraybackslash}X} @{\hspace{3pt}} *{5}{>{\centering\arraybackslash}X}}
    \toprule
     & \multicolumn{5}{c}{\textbf{ALL Data}} & \multicolumn{5}{c}{\textbf{w/o Irrelevant}} & \multicolumn{5}{c}{\textbf{w/o Irrelevant, Outlier}} \\
    \cmidrule(lr){2-6} \cmidrule(lr){7-11} \cmidrule(lr){12-16}
    \textbf{Method} & \textbf{Pos.} & \textbf{Neg.} & \textbf{Neu.} & \textbf{Inc.} & \textbf{Total} & \textbf{Pos.} & \textbf{Neg.} & \textbf{Neu.} & \textbf{Inc.} & \textbf{Total} & \textbf{Pos.} & \textbf{Neg.} & \textbf{Neu.} & \textbf{Inc.} & \textbf{Total} \\ 
    \midrule
    DIRECT & 9605 & 3535 & 1019 & 4561 & 18720 & 8921 & 3203 & 871 & 4337 & 17332 & 7484 & 2701 & 565 & 4011 & 14761 \\
    SSAS & 9688 & 3623 & 951 & 4458 & 18720 & 8973 & 3257 & 833 & 4269 & 17332 & 7430 & 2753 & 587 & 3991 & 14761 \\
    \textbf{Diff} & 83 & 88 & 68 & 103 & 342 & 52 & 54 & 38 & 68 & 212 & 54 & 52 & 22 & 20 & 148 \\ 
    \midrule
    DIRECT \% & 51.31\% & 18.88\% & 5.44\% & 24.36\% & 100\% & 51.47\% & 18.48\% & 5.03\% & 25.02\% & 100\% & 50.70\% & 18.30\% & 3.83\% & 27.17\% & 100\% \\
    SSAS \% & 51.75\% & 19.35\% & 5.08\% & 23.81\% & 100\% & 51.77\% & 18.79\% & 4.81\% & 24.63\% & 100\% & 50.34\% & 18.65\% & 3.98\% & 27.04\% & 100\% \\
    \% Diff & 0.44\% & 0.47\% & 0.36\% & 0.55\% & 1.83\% & 0.30\% & 0.31\% & 0.22\% & 0.39\% & 1.22\% & 0.37\% & 0.35\% & 0.15\% & 0.14\% & 1.00\% \\ 
    \midrule
    \textbf{Improvement} & \multicolumn{2}{c}{\textcolor{blue}{1.83\%}} & \multicolumn{2}{c}{\textcolor{blue}{0.00\%}} & \textcolor{blue}{1.83\%} & \multicolumn{2}{c}{\textcolor{blue}{1.22\%}} & \multicolumn{2}{c}{\textcolor{blue}{7.41\%}} & \textcolor{blue}{8.64\%} & \multicolumn{2}{c}{\textcolor{blue}{1.00\%}} & \multicolumn{2}{c}{\textcolor{blue}{21.15\%}} & \textcolor{blue}{22.15\%} \\
    \bottomrule
    \end{tabularx}
    \vspace{8pt}
    \caption{Detailed Consistency Metrics for Goodreads Book Reviews (Scenario 2).}
\end{table*}

\begin{table}[H]
\centering
\scriptsize
\renewcommand{\arraystretch}{1.3}
\setlength{\tabcolsep}{1pt}

\begin{tabularx}{\textwidth}{l l *{5}{>{\centering\arraybackslash}X} @{\hspace{2pt}} *{5}{>{\centering\arraybackslash}X} @{\hspace{2pt}} *{5}{>{\centering\arraybackslash}X}}
\toprule
 & & \multicolumn{15}{c}{\textbf{SSAS}} \\
 \cmidrule(lr){3-17}
 & & \multicolumn{5}{c}{\textbf{ALL Data}} & \multicolumn{5}{c}{\textbf{w/o Irrelevant}} & \multicolumn{5}{c}{\textbf{w/o Irrelevant, Outlier}} \\
\cmidrule(lr){3-7} \cmidrule(lr){8-12} \cmidrule(lr){13-17}
\textbf{Source} & \textbf{Datapoints} & \textbf{Pos.} & \textbf{Neg.} & \textbf{Neu.} & \textbf{Inc.} & \textbf{Total} & \textbf{Pos.} & \textbf{Neg.} & \textbf{Neu.} & \textbf{Inc.} & \textbf{Total} & \textbf{Pos.} & \textbf{Neg.} & \textbf{Neu.} & \textbf{Inc.} & \textbf{Total} \\
\midrule

\multirow{4}{*}{\rotatebox[origin=c]{90}{\textbf{DIRECT}}} 
& \textbf{Positive} & 9,104 & \textcolor{orange}{16} & \textcolor{orange}{7} & \textcolor{red}{478} & 9,605 & 8,458 & \textcolor{orange}{10} & \textcolor{orange}{5} & \textcolor{red}{448} & 8,921 & 7,084 & \textcolor{orange}{0} & \textcolor{orange}{0} & \textcolor{red}{400} & 7,484 \\

& \textbf{Negative} & \textcolor{orange}{32} & 3,226 & \textcolor{orange}{17} & \textcolor{red}{260} & 3,535 & \textcolor{orange}{22} & 2,920 & \textcolor{orange}{16} & \textcolor{red}{245} & 3,203 & \textcolor{orange}{0} & 2,492 & \textcolor{orange}{0} & \textcolor{red}{209} & 2,701 \\

& \textbf{Neutral} & \textcolor{orange}{57} & \textcolor{orange}{23} & 687 & \textcolor{red}{252} & 1,019 & \textcolor{orange}{48} & \textcolor{orange}{18} & 585 & \textcolor{red}{220} & 871 & \textcolor{orange}{5} & \textcolor{orange}{0} & 394 & \textcolor{red}{166} & 565 \\
\bottomrule
\end{tabularx}

\vspace{8pt}
\caption{Confusion Matrix (Goodreads Book Reviews): Scenario 2 (HIGH Volume, 51--99\% Dist.)}
\label{tab:appx_cm_goodreads_s2}
\end{table}

\clearpage
\section{Scenario 3 – HIGH Volume, 0--50\% Review Distribution}
\subsection{Amazon Product Reviews}

\begin{table*}[h!]
    \centering
    \scriptsize
    \renewcommand{\arraystretch}{1.3}
    \setlength{\tabcolsep}{1.1pt}
    \begin{tabularx}{\textwidth}{l *{5}{>{\centering\arraybackslash}X} @{\hspace{3pt}} *{5}{>{\centering\arraybackslash}X} @{\hspace{3pt}} *{5}{>{\centering\arraybackslash}X}}
    \toprule
     & \multicolumn{5}{c}{\textbf{ALL Data}} & \multicolumn{5}{c}{\textbf{w/o Irrelevant}} & \multicolumn{5}{c}{\textbf{w/o Irrelevant, Outlier}} \\
    \cmidrule(lr){2-6} \cmidrule(lr){7-11} \cmidrule(lr){12-16}
    \textbf{Method} & \textbf{Pos.} & \textbf{Neg.} & \textbf{Neu.} & \textbf{Inc.} & \textbf{Total} & \textbf{Pos.} & \textbf{Neg.} & \textbf{Neu.} & \textbf{Inc.} & \textbf{Total} & \textbf{Pos.} & \textbf{Neg.} & \textbf{Neu.} & \textbf{Inc.} & \textbf{Total} \\ 
    \midrule
    DIRECT & 22,414 & 12,342 & 1,904 & 5,817 & 42,477 & 21,779 & 11,953 & 1,755 & 5,716 & 41,203 & 17,917 & 9,145 & 1,035 & 4,879 & 32,976 \\
    SSAS & 22,881 & 12,568 & 1,874 & 5,154 & 42,477 & 22,209 & 12,176 & 1,736 & 5,082 & 41,203 & 18,133 & 9,285 & 1,065 & 4,493 & 32,976 \\
    \textbf{Diff} & 467 & 226 & 30 & 663 & 1,386 & 430 & 223 & 19 & 634 & 1,306 & 216 & 140 & 30 & 386 & 772 \\ 
    \midrule
    DIRECT \% & 52.77\% & 29.06\% & 4.48\% & 13.69\% & 100\% & 52.86\% & 29.01\% & 4.26\% & 13.87\% & 100\% & 54.33\% & 27.73\% & 3.14\% & 14.80\% & 100\% \\
    SSAS \% & 53.87\% & 29.59\% & 4.41\% & 12.13\% & 100\% & 53.90\% & 29.55\% & 4.21\% & 12.33\% & 100\% & 54.99\% & 28.16\% & 3.23\% & 13.63\% & 100\% \\
    \% Diff & 1.10\% & 0.53\% & 0.07\% & 1.56\% & 3.26\% & 1.04\% & 0.54\% & 0.05\% & 1.54\% & 3.17\% & 0.66\% & 0.42\% & 0.09\% & 1.17\% & 2.34\% \\ 
    \midrule
    \textbf{Improvement} & \multicolumn{2}{c}{\textcolor{blue}{3.26\%}} & \multicolumn{2}{c}{\textcolor{blue}{0.00\%}} & \textcolor{blue}{3.26\%} & \multicolumn{2}{c}{\textcolor{blue}{3.17\%}} & \multicolumn{2}{c}{\textcolor{blue}{3.00\%}} & \textcolor{blue}{6.17\%} & \multicolumn{2}{c}{\textcolor{blue}{2.34\%}} & \multicolumn{2}{c}{\textcolor{blue}{22.37\%}} & \textcolor{blue}{24.71\%} \\
    \bottomrule
    \end{tabularx}
    \vspace{8pt}
    \caption{Detailed Consistency Metrics for Amazon Product Reviews (Scenario 3).}
    \label{tab:appx_amazon_s3}
\end{table*}

\begin{table}[H]
\centering
\scriptsize
\renewcommand{\arraystretch}{1.3}
\setlength{\tabcolsep}{1pt}

\begin{tabularx}{\textwidth}{l l *{5}{>{\centering\arraybackslash}X} @{\hspace{2pt}} *{5}{>{\centering\arraybackslash}X} @{\hspace{2pt}} *{5}{>{\centering\arraybackslash}X}}
\toprule
 & & \multicolumn{15}{c}{\textbf{SSAS}} \\
 \cmidrule(lr){3-17}
 & & \multicolumn{5}{c}{\textbf{ALL Data}} & \multicolumn{5}{c}{\textbf{w/o Irrelevant}} & \multicolumn{5}{c}{\textbf{w/o Irrelevant, Outlier}} \\
\cmidrule(lr){3-7} \cmidrule(lr){8-12} \cmidrule(lr){13-17}
\textbf{Source} & \textbf{Datapoints} & \textbf{Pos.} & \textbf{Neg.} & \textbf{Neu.} & \textbf{Inc.} & \textbf{Total} & \textbf{Pos.} & \textbf{Neg.} & \textbf{Neu.} & \textbf{Inc.} & \textbf{Total} & \textbf{Pos.} & \textbf{Neg.} & \textbf{Neu.} & \textbf{Inc.} & \textbf{Total} \\
\midrule

\multirow{4}{*}{\rotatebox[origin=c]{90}{\textbf{DIRECT}}} 
& \textbf{Positive} & 21,717 & \textcolor{orange}{19} & \textcolor{orange}{39} & \textcolor{red}{639} & 22,414 & 21,105 & \textcolor{orange}{16} & \textcolor{orange}{31} & \textcolor{red}{627} & 21,779 & 17,380 & \textcolor{orange}{0} & \textcolor{orange}{2} & \textcolor{red}{535} & 17,917 \\

& \textbf{Negative} & \textcolor{orange}{31} & 11,894 & \textcolor{orange}{73} & \textcolor{red}{344} & 12,342 & \textcolor{orange}{28} & 11,526 & \textcolor{orange}{66} & \textcolor{red}{333} & 11,953 & \textcolor{orange}{0} & 8,882 & \textcolor{orange}{1} & \textcolor{red}{262} & 9,145 \\

& \textbf{Neutral} & \textcolor{orange}{132} & \textcolor{orange}{57} & 1,321 & \textcolor{red}{394} & 1,904 & \textcolor{orange}{118} & \textcolor{orange}{53} & 1,205 & \textcolor{red}{379} & 1,755 & \textcolor{orange}{21} & \textcolor{orange}{0} & 751 & \textcolor{red}{263} & 1,035 \\
\bottomrule
\end{tabularx}

\vspace{8pt}
\caption{Confusion Matrix (Amazon Product Reviews): Scenario 3 (HIGH Volume, 0--50\% Dist.)}
\label{tab:appx_cm_amazon_s3}
\end{table}

\subsection{Google Business Reviews}

\begin{table*}[h!]
    \centering
    \scriptsize
    \renewcommand{\arraystretch}{1.3}
    \setlength{\tabcolsep}{1.1pt}
    \begin{tabularx}{\textwidth}{l *{5}{>{\centering\arraybackslash}X} @{\hspace{3pt}} *{5}{>{\centering\arraybackslash}X} @{\hspace{3pt}} *{5}{>{\centering\arraybackslash}X}}
    \toprule
     & \multicolumn{5}{c}{\textbf{ALL Data}} & \multicolumn{5}{c}{\textbf{w/o Irrelevant}} & \multicolumn{5}{c}{\textbf{w/o Irrelevant, Outlier}} \\
    \cmidrule(lr){2-6} \cmidrule(lr){7-11} \cmidrule(lr){12-16}
    \textbf{Method} & \textbf{Pos.} & \textbf{Neg.} & \textbf{Neu.} & \textbf{Inc.} & \textbf{Total} & \textbf{Pos.} & \textbf{Neg.} & \textbf{Neu.} & \textbf{Inc.} & \textbf{Total} & \textbf{Pos.} & \textbf{Neg.} & \textbf{Neu.} & \textbf{Inc.} & \textbf{Total} \\ 
    \midrule
    DIRECT & 4,016 & 780 & 156 & 566 & 5,518 & 3,911 & 756 & 146 & 555 & 5,368 & 3,512 & 588 & 86 & 433 & 4,619 \\
    SSAS & 4,031 & 826 & 165 & 496 & 5,518 & 3,924 & 801 & 152 & 491 & 5,368 & 3,504 & 612 & 92 & 411 & 4,619 \\
    \textbf{Diff} & 15 & 46 & 9 & --70 & 140 & 13 & 45 & 6 & --64 & 128 & --8 & 24 & 6 & --22 & 60 \\ 
    \midrule
    DIRECT \% & 72.78\% & 14.14\% & 2.83\% & 10.26\% & 100\% & 72.86\% & 14.08\% & 2.72\% & 10.34\% & 100\% & 76.03\% & 12.73\% & 1.86\% & 9.37\% & 100\% \\
    SSAS \% & 73.05\% & 14.97\% & 2.99\% & 8.99\% & 100\% & 73.10\% & 14.92\% & 2.83\% & 9.15\% & 100\% & 75.86\% & 13.25\% & 1.99\% & 8.90\% & 100\% \\
    \% Diff & 0.27\% & 0.83\% & 0.16\% & --1.27\% & 2.54\% & 0.24\% & 0.84\% & 0.11\% & --1.19\% & 2.38\% & --0.17\% & 0.52\% & 0.13\% & --0.48\% & 1.30\% \\ 
    \midrule
    \textbf{Improvement} & \multicolumn{2}{c}{\textcolor{blue}{2.54\%}} & \multicolumn{2}{c}{\textcolor{blue}{0.00\%}} & \textcolor{blue}{2.54\%} & \multicolumn{2}{c}{\textcolor{blue}{2.38\%}} & \multicolumn{2}{c}{\textcolor{blue}{2.72\%}} & \textcolor{blue}{5.10\%} & \multicolumn{2}{c}{\textcolor{blue}{1.30\%}} & \multicolumn{2}{c}{\textcolor{blue}{16.29\%}} & \textcolor{blue}{17.59\%} \\
    \bottomrule
    \end{tabularx}
    \vspace{8pt}
    \caption{Detailed Consistency Metrics for Google Business Reviews (Scenario 3).}
\end{table*}

\begin{table}[H]
\centering
\scriptsize
\renewcommand{\arraystretch}{1.3}
\setlength{\tabcolsep}{1pt}

\begin{tabularx}{\textwidth}{l l *{5}{>{\centering\arraybackslash}X} @{\hspace{2pt}} *{5}{>{\centering\arraybackslash}X} @{\hspace{2pt}} *{5}{>{\centering\arraybackslash}X}}
\toprule
 & & \multicolumn{15}{c}{\textbf{SSAS}} \\
 \cmidrule(lr){3-17}
 & & \multicolumn{5}{c}{\textbf{ALL Data}} & \multicolumn{5}{c}{\textbf{w/o Irrelevant}} & \multicolumn{5}{c}{\textbf{w/o Irrelevant, Outlier}} \\
\cmidrule(lr){3-7} \cmidrule(lr){8-12} \cmidrule(lr){13-17}
\textbf{Source} & \textbf{Datapoints} & \textbf{Pos.} & \textbf{Neg.} & \textbf{Neu.} & \textbf{Inc.} & \textbf{Total} & \textbf{Pos.} & \textbf{Neg.} & \textbf{Neu.} & \textbf{Inc.} & \textbf{Total} & \textbf{Pos.} & \textbf{Neg.} & \textbf{Neu.} & \textbf{Inc.} & \textbf{Total} \\
\midrule

\multirow{4}{*}{\rotatebox[origin=c]{90}{\textbf{DIRECT}}} 
& \textbf{Positive} & 3,942 & \textcolor{orange}{1} & \textcolor{orange}{6} & \textcolor{red}{67} & 4,016 & 3,839 & \textcolor{orange}{1} & \textcolor{orange}{6} & \textcolor{red}{65} & 3,911 & 3,459 & \textcolor{orange}{0} & \textcolor{orange}{0} & \textcolor{red}{53} & 3,512 \\

& \textbf{Negative} & \textcolor{orange}{2} & 751 & \textcolor{orange}{4} & \textcolor{red}{23} & 780 & \textcolor{orange}{2} & 728 & \textcolor{orange}{3} & \textcolor{red}{23} & 756 & \textcolor{orange}{0} & 573 & \textcolor{orange}{0} & \textcolor{red}{15} & 588 \\

& \textbf{Neutral} & \textcolor{orange}{8} & \textcolor{orange}{3} & 112 & \textcolor{red}{33} & 156 & \textcolor{orange}{8} & \textcolor{orange}{3} & 102 & \textcolor{red}{33} & 146 & \textcolor{orange}{2} & \textcolor{orange}{0} & 66 & \textcolor{red}{18} & 86 \\

& \textbf{Incon.} & \textcolor{green!60!black}{79} & \textcolor{green!60!black}{71} & \textcolor{green!60!black}{43} & 373 & 566 & \textcolor{green!60!black}{75} & \textcolor{green!60!black}{69} & \textcolor{green!60!black}{41} & 370 & 555 & \textcolor{green!60!black}{43} & \textcolor{green!60!black}{39} & \textcolor{green!60!black}{26} & 325 & 433 \\
\bottomrule
\end{tabularx}

\vspace{8pt}
\caption{Confusion Matrix (Google Business Reviews): Scenario 3 (HIGH Volume, 0--50\% Dist.)}
\label{tab:appx_cm_google_s3}
\end{table}

\subsection{Goodreads Book Reviews}

\begin{table*}[h!]
    \centering
    \scriptsize
    \renewcommand{\arraystretch}{1.3}
    \setlength{\tabcolsep}{1.1pt}
    \begin{tabularx}{\textwidth}{l *{5}{>{\centering\arraybackslash}X} @{\hspace{3pt}} *{5}{>{\centering\arraybackslash}X} @{\hspace{3pt}} *{5}{>{\centering\arraybackslash}X}}
    \toprule
     & \multicolumn{5}{c}{\textbf{ALL Data}} & \multicolumn{5}{c}{\textbf{w/o Irrelevant}} & \multicolumn{5}{c}{\textbf{w/o Irrelevant, Outlier}} \\
    \cmidrule(lr){2-6} \cmidrule(lr){7-11} \cmidrule(lr){12-16}
    \textbf{Method} & \textbf{Pos.} & \textbf{Neg.} & \textbf{Neu.} & \textbf{Inc.} & \textbf{Total} & \textbf{Pos.} & \textbf{Neg.} & \textbf{Neu.} & \textbf{Inc.} & \textbf{Total} & \textbf{Pos.} & \textbf{Neg.} & \textbf{Neu.} & \textbf{Inc.} & \textbf{Total} \\ 
    \midrule
    DIRECT & 49670 & 17716 & 4838 & 20989 & 93213 & 45491 & 15713 & 3974 & 19584 & 84762 & \#VALUE! & \#VALUE! & \#VALUE! & \#VALUE! & \#VALUE! \\
    SSAS & 50223 & 18029 & 4145 & 20816 & 93213 & 45794 & 15923 & 3438 & 19607 & 84762 & \#VALUE! & \#VALUE! & \#VALUE! & \#VALUE! & \#VALUE! \\
    \textbf{Diff} & 553 & 313 & 693 & 173 & 1732 & 303 & 210 & 536 & 23 & 1072 & \#VALUE! & \#VALUE! & \#VALUE! & \#VALUE! & \#VALUE! \\ 
    \midrule
    DIRECT \% & 53.29\% & 19.01\% & 5.19\% & 22.52\% & 100\% & 53.67\% & 18.54\% & 4.69\% & 23.10\% & 100\% & \#VALUE! & \#VALUE! & \#VALUE! & \#VALUE! & \#VALUE! \\
    SSAS \% & 53.88\% & 19.34\% & 4.45\% & 22.33\% & 100\% & 54.03\% & 18.79\% & 4.06\% & 23.13\% & 100\% & \#VALUE! & \#VALUE! & \#VALUE! & \#VALUE! & \#VALUE! \\
    \% Diff & 0.59\% & 0.34\% & 0.74\% & 0.19\% & 1.86\% & 0.36\% & 0.25\% & 0.63\% & 0.03\% & 1.26\% & \#VALUE! & \#VALUE! & \#VALUE! & \#VALUE! & \#VALUE! \\
\bottomrule
    \end{tabularx}
    \vspace{8pt}
    \caption{Detailed Consistency Metrics for Goodreads Book Reviews (Scenario 3).}
\end{table*}

\begin{table}[H]
\centering
\scriptsize
\renewcommand{\arraystretch}{1.3}
\setlength{\tabcolsep}{1pt}

\begin{tabularx}{\textwidth}{l l *{5}{>{\centering\arraybackslash}X} @{\hspace{2pt}} *{5}{>{\centering\arraybackslash}X} @{\hspace{2pt}} *{5}{>{\centering\arraybackslash}X}}
\toprule
 & & \multicolumn{15}{c}{\textbf{SSAS}} \\
 \cmidrule(lr){3-17}
 & & \multicolumn{5}{c}{\textbf{ALL Data}} & \multicolumn{5}{c}{\textbf{w/o Irrelevant}} & \multicolumn{5}{c}{\textbf{w/o Irrelevant, Outlier}} \\
\cmidrule(lr){3-7} \cmidrule(lr){8-12} \cmidrule(lr){13-17}
\textbf{Source} & \textbf{Datapoints} & \textbf{Pos.} & \textbf{Neg.} & \textbf{Neu.} & \textbf{Inc.} & \textbf{Total} & \textbf{Pos.} & \textbf{Neg.} & \textbf{Neu.} & \textbf{Inc.} & \textbf{Total} & \textbf{Pos.} & \textbf{Neg.} & \textbf{Neu.} & \textbf{Inc.} & \textbf{Total} \\
\midrule

\multirow{4}{*}{\rotatebox[origin=c]{90}{\textbf{DIRECT}}} 
& \textbf{Positive} & 47,082 & \textcolor{orange}{80} & \textcolor{orange}{67} & \textcolor{red}{2,441} & 49,670 & 43,103 & \textcolor{orange}{59} & \textcolor{orange}{38} & \textcolor{red}{2,291} & 45,491 & 34,428 & \textcolor{orange}{0} & \textcolor{orange}{1} & \textcolor{red}{2,067} & 36,496 \\

& \textbf{Negative} & \textcolor{orange}{196} & 16,109 & \textcolor{orange}{75} & \textcolor{red}{1,336} & 17,716 & \textcolor{orange}{129} & 14,288 & \textcolor{orange}{50} & \textcolor{red}{1,246} & 15,713 & \textcolor{orange}{0} & 11,896 & \textcolor{orange}{0} & \textcolor{red}{1,056} & 12,952 \\

& \textbf{Neutral} & \textcolor{orange}{342} & \textcolor{orange}{152} & 2,983 & \textcolor{red}{1,361} & 4,838 & \textcolor{orange}{265} & \textcolor{orange}{108} & 2,436 & \textcolor{red}{1,165} & 3,974 & \textcolor{orange}{20} & \textcolor{orange}{2} & 1,596 & \textcolor{red}{879} & 2,497 \\
\bottomrule
\end{tabularx}

\vspace{8pt}
\caption{Confusion Matrix (Goodreads Book Reviews): Scenario 3 (HIGH Volume, 0--50\% Dist.)}
\label{tab:appx_cm_goodreads_s3}
\end{table}

\clearpage
\section{Scenario 4 – LOW Volume, 100\% Review Distribution}

\subsection{Amazon Product Reviews}
No Data for this combination of Dataset and Scenario

\subsection{Google Business Reviews}
No Data for this combination of Dataset and Scenario

\subsection{Goodreads Book Reviews}
No Data for this combination of Dataset and Scenario

\clearpage
\section{Scenario 5 – LOW Volume, 51--99\% Review Distribution}
\subsection{Amazon Product Reviews}

\begin{table*}[h!]
    \centering
    \scriptsize
    \renewcommand{\arraystretch}{1.3}
    \setlength{\tabcolsep}{1.1pt}
\begin{tabularx}{\textwidth}{l *{5}{>{\centering\arraybackslash}X} @{\hspace{3pt}} *{5}{>{\centering\arraybackslash}X} @{\hspace{3pt}} *{5}{>{\centering\arraybackslash}X}}
    \toprule
     & \multicolumn{5}{c}{\textbf{ALL Data}} & \multicolumn{5}{c}{\textbf{w/o Irrelevant}} & \multicolumn{5}{c}{\textbf{w/o Irrelevant, Outlier}} \\
    \cmidrule(lr){2-6} \cmidrule(lr){7-11} \cmidrule(lr){12-16}
    \textbf{Method} & \textbf{Pos.} & \textbf{Neg.} & \textbf{Neu.} & \textbf{Inc.} & \textbf{Total} & \textbf{Pos.} & \textbf{Neg.} & \textbf{Neu.} & \textbf{Inc.} & \textbf{Total} & \textbf{Pos.} & \textbf{Neg.} & \textbf{Neu.} & \textbf{Inc.} & \textbf{Total} \\ 
    \midrule
    DIRECT & 27 & 4 & 5 & 8 & 44 & 26 & 4 & 4 & 7 & 41 & 19 & 3 & 2 & 7 & 31 \\
    SSAS & 28 & 4 & 5 & 7 & 44 & 26 & 4 & 4 & 7 & 41 & 18 & 3 & 3 & 7 & 31 \\
    \textbf{Diff} & 1 & 0 & 0 & 1 & 2 & 0 & 0 & 0 & 0 & 0 & 1 & 0 & 1 & 0 & 2 \\ 
    \midrule
    DIRECT \% & 61.36\% & 9.09\% & 11.36\% & 18.18\% & 100\% & 63.41\% & 9.76\% & 9.76\% & 17.07\% & 100\% & 61.29\% & 9.68\% & 6.45\% & 22.58\% & 100\% \\
    SSAS \% & 63.64\% & 9.09\% & 11.36\% & 15.91\% & 100\% & 63.41\% & 9.76\% & 9.76\% & 17.07\% & 100\% & 58.06\% & 9.68\% & 9.68\% & 22.58\% & 100\% \\
    \% Diff & 2.27\% & 0.00\% & 0.00\% & 2.27\% & 4.55\% & 0.00\% & 0.00\% & 0.00\% & 0.00\% & 0.00\% & 3.23\% & 0.00\% & 3.23\% & 0.00\% & 6.45\% \\ 
    \midrule
    \textbf{Improvement} & \multicolumn{2}{c}{\textcolor{blue}{4.55\%}} & \multicolumn{2}{c}{\textcolor{blue}{0.00\%}} & \textcolor{blue}{4.55\%} & \multicolumn{2}{c}{\textcolor{blue}{0.00\%}} & \multicolumn{2}{c}{\textcolor{blue}{6.82\%}} & \textcolor{blue}{6.82\%} & \multicolumn{2}{c}{\textcolor{blue}{6.45\%}} & \multicolumn{2}{c}{\textcolor{blue}{29.55\%}} & \textcolor{blue}{36.00\%} \\
    \bottomrule
    \end{tabularx}
    \vspace{8pt}
    \caption{Detailed Consistency Metrics for Amazon Product Reviews (Scenario 5).}
    \label{tab:appx_amazon_s5}
\end{table*}

\begin{table}[H]
\centering
\scriptsize
\renewcommand{\arraystretch}{1.3}
\setlength{\tabcolsep}{1pt}

\begin{tabularx}{\textwidth}{l l *{5}{>{\centering\arraybackslash}X} @{\hspace{2pt}} *{5}{>{\centering\arraybackslash}X} @{\hspace{2pt}} *{5}{>{\centering\arraybackslash}X}}
\toprule
 & & \multicolumn{15}{c}{\textbf{SSAS}} \\
 \cmidrule(lr){3-17}
 & & \multicolumn{5}{c}{\textbf{ALL Data}} & \multicolumn{5}{c}{\textbf{w/o Irrelevant}} & \multicolumn{5}{c}{\textbf{w/o Irrelevant, Outlier}} \\
\cmidrule(lr){3-7} \cmidrule(lr){8-12} \cmidrule(lr){13-17}
\textbf{Source} & \textbf{Datapoints} & \textbf{Pos.} & \textbf{Neg.} & \textbf{Neu.} & \textbf{Inc.} & \textbf{Total} & \textbf{Pos.} & \textbf{Neg.} & \textbf{Neu.} & \textbf{Inc.} & \textbf{Total} & \textbf{Pos.} & \textbf{Neg.} & \textbf{Neu.} & \textbf{Inc.} & \textbf{Total} \\
\midrule

\multirow{4}{*}{\rotatebox[origin=c]{90}{\textbf{DIRECT}}} 
& \textbf{Positive} & 25 & \textcolor{orange}{0} & \textcolor{orange}{0} & \textcolor{red}{2} & 27 & 24 & \textcolor{orange}{0} & \textcolor{orange}{0} & \textcolor{red}{2} & 26 & 17 & \textcolor{orange}{0} & \textcolor{orange}{0} & \textcolor{red}{2} & 19 \\

& \textbf{Negative} & \textcolor{orange}{0} & 3 & \textcolor{orange}{0} & \textcolor{red}{1} & 4 & \textcolor{orange}{0} & 3 & \textcolor{orange}{0} & \textcolor{red}{1} & 4 & \textcolor{orange}{0} & 2 & \textcolor{orange}{0} & \textcolor{red}{1} & 3 \\

& \textbf{Neutral} & \textcolor{orange}{1} & \textcolor{orange}{0} & 4 & \textcolor{red}{0} & 5 & \textcolor{orange}{1} & \textcolor{orange}{0} & 3 & \textcolor{red}{0} & 4 & \textcolor{orange}{0} & \textcolor{orange}{0} & 2 & \textcolor{red}{0} & 2 \\

& \textbf{Incon.} & \textcolor{green!60!black}{2} & \textcolor{green!60!black}{1} & \textcolor{green!60!black}{1} & 4 & 8 & \textcolor{green!60!black}{1} & \textcolor{green!60!black}{1} & \textcolor{green!60!black}{1} & 4 & 7 & \textcolor{green!60!black}{1} & \textcolor{green!60!black}{1} & \textcolor{green!60!black}{1} & 4 & 7 \\
\midrule
& \textbf{Total} & 28 & 4 & 5 & 7 & & 26 & 4 & 4 & 7 & & 18 & 3 & 3 & 7 & \\
\bottomrule
\end{tabularx}

\vspace{8pt}
\caption{Confusion Matrix (Amazon Product Reviews): Scenario 5 (LOW Volume, 51--99\% Dist.)}
\label{tab:appx_cm_amazon_s5}
\end{table}

\subsection{Google Business Reviews}

\begin{table*}[h!]
    \centering
    \scriptsize
    \renewcommand{\arraystretch}{1.3}
    \setlength{\tabcolsep}{1.1pt}
    \begin{tabularx}{\textwidth}{l *{5}{>{\centering\arraybackslash}X} @{\hspace{3pt}} *{5}{>{\centering\arraybackslash}X} @{\hspace{3pt}} *{5}{>{\centering\arraybackslash}X}}
    \toprule
     & \multicolumn{5}{c}{\textbf{ALL Data}} & \multicolumn{5}{c}{\textbf{w/o Irrelevant}} & \multicolumn{5}{c}{\textbf{w/o Irrelevant, Outlier}} \\
    \cmidrule(lr){2-6} \cmidrule(lr){7-11} \cmidrule(lr){12-16}
    \textbf{Method} & \textbf{Pos.} & \textbf{Neg.} & \textbf{Neu.} & \textbf{Inc.} & \textbf{Total} & \textbf{Pos.} & \textbf{Neg.} & \textbf{Neu.} & \textbf{Inc.} & \textbf{Total} & \textbf{Pos.} & \textbf{Neg.} & \textbf{Neu.} & \textbf{Inc.} & \textbf{Total} \\ 
    \midrule
    DIRECT & 6,041 & 1,566 & 324 & 803 & 8,734 & 5,860 & 1,509 & 277 & 767 & 8,413 & 5,139 & 1,162 & 148 & 606 & 7,055 \\
    SSAS & 6,070 & 1,604 & 305 & 755 & 8,734 & 5,881 & 1,540 & 261 & 731 & 8,413 & 5,120 & 1,187 & 142 & 606 & 7,055 \\
    \textbf{Diff} & 29 & 38 & --19 & --48 & 134 & 21 & 31 & --16 & --36 & 104 & --19 & 25 & --6 & 0 & 50 \\ 
    \midrule
    DIRECT \% & 69.17\% & 17.93\% & 3.71\% & 9.19\% & 100\% & 69.65\% & 17.94\% & 3.29\% & 9.12\% & 100\% & 72.84\% & 16.47\% & 2.10\% & 8.59\% & 100\% \\
    SSAS \% & 69.50\% & 18.37\% & 3.49\% & 8.64\% & 100\% & 69.90\% & 18.31\% & 3.10\% & 8.69\% & 100\% & 72.57\% & 16.82\% & 2.01\% & 8.59\% & 100\% \\
    \% Diff & 0.33\% & 0.44\% & --0.22\% & --0.55\% & 1.53\% & 0.25\% & 0.37\% & --0.19\% & --0.43\% & 1.24\% & --0.27\% & 0.35\% & --0.09\% & 0.00\% & 0.71\% \\ 
    \midrule
    \textbf{Improvement} & \multicolumn{2}{c}{\textcolor{blue}{1.53\%}} & \multicolumn{2}{c}{\textcolor{blue}{0.00\%}} & \textcolor{blue}{1.53\%} & \multicolumn{2}{c}{\textcolor{blue}{1.24\%}} & \multicolumn{2}{c}{\textcolor{blue}{3.68\%}} & \textcolor{blue}{4.91\%} & \multicolumn{2}{c}{\textcolor{blue}{0.71\%}} & \multicolumn{2}{c}{\textcolor{blue}{19.22\%}} & \textcolor{blue}{19.93\%} \\
    \bottomrule
    \end{tabularx}
    \vspace{8pt}
    \caption{Detailed Consistency Metrics for Google Business Reviews (Scenario 5).}
\end{table*}

\begin{table}[H]
\centering
\scriptsize
\renewcommand{\arraystretch}{1.3}
\setlength{\tabcolsep}{1pt}

\begin{tabularx}{\textwidth}{l l *{5}{>{\centering\arraybackslash}X} @{\hspace{2pt}} *{5}{>{\centering\arraybackslash}X} @{\hspace{2pt}} *{5}{>{\centering\arraybackslash}X}}
\toprule
 & & \multicolumn{15}{c}{\textbf{SSAS}} \\
 \cmidrule(lr){3-17}
 & & \multicolumn{5}{c}{\textbf{ALL Data}} & \multicolumn{5}{c}{\textbf{w/o Irrelevant}} & \multicolumn{5}{c}{\textbf{w/o Irrelevant, Outlier}} \\
\cmidrule(lr){3-7} \cmidrule(lr){8-12} \cmidrule(lr){13-17}
\textbf{Source} & \textbf{Datapoints} & \textbf{Pos.} & \textbf{Neg.} & \textbf{Neu.} & \textbf{Inc.} & \textbf{Total} & \textbf{Pos.} & \textbf{Neg.} & \textbf{Neu.} & \textbf{Inc.} & \textbf{Total} & \textbf{Pos.} & \textbf{Neg.} & \textbf{Neu.} & \textbf{Inc.} & \textbf{Total} \\
\midrule

\multirow{4}{*}{\rotatebox[origin=c]{90}{\textbf{DIRECT}}} 
& \textbf{Positive} & 5,914 & \textcolor{orange}{4} & \textcolor{orange}{13} & \textcolor{red}{110} & 6,041 & 5,737 & \textcolor{orange}{3} & \textcolor{orange}{12} & \textcolor{red}{108} & 5,860 & 5,058 & \textcolor{orange}{0} & \textcolor{orange}{0} & \textcolor{red}{81} & 5,139 \\

& \textbf{Negative} & \textcolor{orange}{3} & 1,515 & \textcolor{orange}{9} & \textcolor{red}{39} & 1,566 & \textcolor{orange}{3} & 1,462 & \textcolor{orange}{7} & \textcolor{red}{37} & 1,509 & \textcolor{orange}{0} & 1,134 & \textcolor{orange}{0} & \textcolor{red}{28} & 1,162 \\

& \textbf{Neutral} & \textcolor{orange}{27} & \textcolor{orange}{4} & 234 & \textcolor{red}{59} & 324 & \textcolor{orange}{22} & \textcolor{orange}{3} & 198 & \textcolor{red}{54} & 277 & \textcolor{orange}{1} & \textcolor{orange}{0} & 117 & \textcolor{red}{30} & 148 \\

& \textbf{Incon.} & \textcolor{green!60!black}{126} & \textcolor{green!60!black}{81} & \textcolor{green!60!black}{49} & 547 & 803 & \textcolor{green!60!black}{119} & \textcolor{green!60!black}{72} & \textcolor{green!60!black}{44} & 532 & 767 & \textcolor{green!60!black}{61} & \textcolor{green!60!black}{53} & \textcolor{green!60!black}{25} & 467 & 606 \\
\bottomrule
\end{tabularx}

\vspace{8pt}
\caption{Confusion Matrix (Google Business Reviews): Scenario 5 (LOW Volume, 51--99\% Dist.)}
\label{tab:appx_cm_google_s5}
\end{table}

\subsection{Goodreads Book Reviews}
No Data for this combination of Dataset and Scenario

\clearpage
\section{Scenario 6 – LOW Volume, 0--50\% Review Distribution}
\subsection{Amazon Product Reviews}

\begin{table*}[h!]
    \centering
    \scriptsize
    \renewcommand{\arraystretch}{1.3}
    \setlength{\tabcolsep}{1.1pt}
    \begin{tabularx}{\textwidth}{l *{5}{>{\centering\arraybackslash}X} @{\hspace{3pt}} *{5}{>{\centering\arraybackslash}X} @{\hspace{3pt}} *{5}{>{\centering\arraybackslash}X}}
    \toprule
     & \multicolumn{5}{c}{\textbf{ALL Data}} & \multicolumn{5}{c}{\textbf{w/o Irrelevant}} & \multicolumn{5}{c}{\textbf{w/o Irrelevant, Outlier}} \\
    \cmidrule(lr){2-6} \cmidrule(lr){7-11} \cmidrule(lr){12-16}
    \textbf{Method} & \textbf{Pos.} & \textbf{Neg.} & \textbf{Neu.} & \textbf{Inc.} & \textbf{Total} & \textbf{Pos.} & \textbf{Neg.} & \textbf{Neu.} & \textbf{Inc.} & \textbf{Total} & \textbf{Pos.} & \textbf{Neg.} & \textbf{Neu.} & \textbf{Inc.} & \textbf{Total} \\ 
    \midrule
    DIRECT & 18,779 & 10,504 & 1,819 & 4,423 & 35,525 & 17,988 & 10,064 & 1,669 & 4,324 & 34,045 & 13,792 & 7,295 & 885 & 3,511 & 25,483 \\
    SSAS & 19,144 & 10,706 & 1,660 & 4,015 & 35,525 & 18,317 & 10,262 & 1,517 & 3,949 & 34,045 & 13,919 & 7,415 & 823 & 3,326 & 25,483 \\
    \textbf{Diff} & 365 & 202 & 159 & 408 & 1,134 & 329 & 198 & 152 & 375 & 1,054 & 127 & 120 & 62 & 185 & 494 \\ 
    \midrule
    DIRECT \% & 52.86\% & 29.57\% & 5.12\% & 12.45\% & 100\% & 52.84\% & 29.56\% & 4.90\% & 12.70\% & 100\% & 54.12\% & 28.63\% & 3.47\% & 13.78\% & 100\% \\
    SSAS \% & 53.89\% & 30.14\% & 4.67\% & 11.30\% & 100\% & 53.80\% & 30.14\% & 4.46\% & 11.60\% & 100\% & 54.62\% & 29.10\% & 3.23\% & 13.05\% & 100\% \\
    \% Diff & 1.03\% & 0.57\% & 0.45\% & 1.15\% & 3.19\% & 0.97\% & 0.58\% & 0.45\% & 1.10\% & 3.10\% & 0.50\% & 0.47\% & 0.24\% & 0.73\% & 1.94\% \\ 
    \midrule
    \textbf{Improvement} & \multicolumn{2}{c}{\textcolor{blue}{3.19\%}} & \multicolumn{2}{c}{\textcolor{blue}{0.00\%}} & \textcolor{blue}{3.19\%} & \multicolumn{2}{c}{\textcolor{blue}{3.10\%}} & \multicolumn{2}{c}{\textcolor{blue}{4.17\%}} & \textcolor{blue}{7.26\%} & \multicolumn{2}{c}{\textcolor{blue}{1.94\%}} & \multicolumn{2}{c}{\textcolor{blue}{28.27\%}} & \textcolor{blue}{30.21\%} \\
    \bottomrule
    \end{tabularx}
    \vspace{8pt}
    \caption{Detailed Consistency Metrics for Amazon Product Reviews (Scenario 6).}
    \label{tab:appx_amazon_s6}
\end{table*}

\begin{table}[H]
\centering
\scriptsize
\renewcommand{\arraystretch}{1.3}
\setlength{\tabcolsep}{1pt}

\begin{tabularx}{\textwidth}{l l *{5}{>{\centering\arraybackslash}X} @{\hspace{2pt}} *{5}{>{\centering\arraybackslash}X} @{\hspace{2pt}} *{5}{>{\centering\arraybackslash}X}}
\toprule
 & & \multicolumn{15}{c}{\textbf{SSAS}} \\
 \cmidrule(lr){3-17}
 & & \multicolumn{5}{c}{\textbf{ALL Data}} & \multicolumn{5}{c}{\textbf{w/o Irrelevant}} & \multicolumn{5}{c}{\textbf{w/o Irrelevant, Outlier}} \\
\cmidrule(lr){3-7} \cmidrule(lr){8-12} \cmidrule(lr){13-17}
\textbf{Source} & \textbf{Datapoints} & \textbf{Pos.} & \textbf{Neg.} & \textbf{Neu.} & \textbf{Inc.} & \textbf{Total} & \textbf{Pos.} & \textbf{Neg.} & \textbf{Neu.} & \textbf{Inc.} & \textbf{Total} & \textbf{Pos.} & \textbf{Neg.} & \textbf{Neu.} & \textbf{Inc.} & \textbf{Total} \\
\midrule

\multirow{4}{*}{\rotatebox[origin=c]{90}{\textbf{DIRECT}}} 
& \textbf{Positive} & 18,131 & \textcolor{orange}{21} & \textcolor{orange}{50} & \textcolor{red}{577} & 18,779 & 17,366 & \textcolor{orange}{20} & \textcolor{orange}{40} & \textcolor{red}{562} & 17,988 & 13,335 & \textcolor{orange}{0} & \textcolor{orange}{4} & \textcolor{red}{453} & 13,792 \\

& \textbf{Negative} & \textcolor{orange}{27} & 10,127 & \textcolor{orange}{78} & \textcolor{red}{272} & 10,504 & \textcolor{orange}{24} & 9,709 & \textcolor{orange}{65} & \textcolor{red}{266} & 10,064 & \textcolor{orange}{0} & 7,107 & \textcolor{orange}{1} & \textcolor{red}{187} & 7,295 \\

& \textbf{Neutral} & \textcolor{orange}{146} & \textcolor{orange}{67} & 1,240 & \textcolor{red}{366} & 1,819 & \textcolor{orange}{126} & \textcolor{orange}{60} & 1,129 & \textcolor{red}{354} & 1,669 & \textcolor{orange}{25} & \textcolor{orange}{0} & 625 & \textcolor{red}{235} & 885 \\
\bottomrule
\end{tabularx}

\vspace{8pt}
\caption{Confusion Matrix (Amazon Product Reviews): Scenario 6 (LOW Volume, 0--50\% Dist.)}
\label{tab:appx_cm_amazon_s6}
\end{table}

\subsection{Google Business Reviews}

\begin{table*}[h!]
    \centering
    \scriptsize
    \renewcommand{\arraystretch}{1.3}
    \setlength{\tabcolsep}{1.1pt}
    \begin{tabularx}{\textwidth}{l *{5}{>{\centering\arraybackslash}X} @{\hspace{3pt}} *{5}{>{\centering\arraybackslash}X} @{\hspace{3pt}} *{5}{>{\centering\arraybackslash}X}}
    \toprule
     & \multicolumn{5}{c}{\textbf{ALL Data}} & \multicolumn{5}{c}{\textbf{w/o Irrelevant}} & \multicolumn{5}{c}{\textbf{w/o Irrelevant, Outlier}} \\
    \cmidrule(lr){2-6} \cmidrule(lr){7-11} \cmidrule(lr){12-16}
    \textbf{Method} & \textbf{Pos.} & \textbf{Neg.} & \textbf{Neu.} & \textbf{Inc.} & \textbf{Total} & \textbf{Pos.} & \textbf{Neg.} & \textbf{Neu.} & \textbf{Inc.} & \textbf{Total} & \textbf{Pos.} & \textbf{Neg.} & \textbf{Neu.} & \textbf{Inc.} & \textbf{Total} \\ 
    \midrule
    DIRECT & 10,014 & 2,175 & 476 & 1,269 & 13,934 & 9,664 & 2,047 & 423 & 1,235 & 13,369 & 8,462 & 1,497 & 213 & 976 & 11,148 \\
    SSAS & 10,067 & 2,224 & 463 & 1,180 & 13,934 & 9,708 & 2,099 & 412 & 1,150 & 13,369 & 8,448 & 1,514 & 224 & 962 & 11,148 \\
    \textbf{Diff} & 53 & 49 & --13 & --89 & 204 & 44 & 52 & --11 & --85 & 192 & --14 & 17 & 11 & --14 & 56 \\ 
    \midrule
    DIRECT \% & 71.87\% & 15.61\% & 3.42\% & 9.11\% & 100\% & 72.29\% & 15.31\% & 3.16\% & 9.24\% & 100\% & 75.91\% & 13.43\% & 1.91\% & 8.75\% & 100\% \\
    SSAS \% & 72.25\% & 15.96\% & 3.32\% & 8.47\% & 100\% & 72.62\% & 15.70\% & 3.08\% & 8.60\% & 100\% & 75.78\% & 13.58\% & 2.01\% & 8.63\% & 100\% \\
    \% Diff & 0.38\% & 0.35\% & --0.09\% & --0.64\% & 1.46\% & 0.33\% & 0.39\% & --0.08\% & --0.64\% & 1.44\% & --0.13\% & 0.15\% & 0.10\% & --0.13\% & 0.50\% \\ 
    \midrule
    \textbf{Improvement} & \multicolumn{2}{c}{\textcolor{blue}{1.46\%}} & \multicolumn{2}{c}{\textcolor{blue}{0.00\%}} & \textcolor{blue}{1.46\%} & \multicolumn{2}{c}{\textcolor{blue}{1.44\%}} & \multicolumn{2}{c}{\textcolor{blue}{4.05\%}} & \textcolor{blue}{5.49\%} & \multicolumn{2}{c}{\textcolor{blue}{0.50\%}} & \multicolumn{2}{c}{\textcolor{blue}{19.99\%}} & \textcolor{blue}{20.50\%} \\
    \bottomrule
    \end{tabularx}
    \vspace{8pt}
    \caption{Detailed Consistency Metrics for Google Business Reviews (Scenario 6).}
\end{table*}

\begin{table}[H]
\centering
\scriptsize
\renewcommand{\arraystretch}{1.3}
\setlength{\tabcolsep}{1pt}

%

\vspace{8pt}
\caption{Confusion Matrix (Google Business Reviews): Scenario 6 (LOW Volume, 0--50\% Dist.)}
\label{tab:appx_cm_google_s6}
\end{table}

\subsection{Goodreads Book Reviews}

\begin{table*}[h!]
    \centering
    \scriptsize
    \renewcommand{\arraystretch}{1.3}
    \setlength{\tabcolsep}{1.1pt}
    %
    \vspace{8pt}
    \caption{Detailed Consistency Metrics for Goodreads Book Reviews (Scenario 6).}
\end{table*}

\begin{table}[H]
\centering
\scriptsize
\renewcommand{\arraystretch}{1.3}
\setlength{\tabcolsep}{1pt}

%

\vspace{8pt}
\caption{Confusion Matrix (Goodreads Book Reviews): Scenario 6 (LOW Volume, 0--50\% Dist.)}
\label{tab:appx_cm_goodreads_s6}
\end{table}

\end{document}